\RequirePackage{fix-cm}
\documentclass[smallextended]{svjour3}       
\smartqed  
\usepackage{graphicx}
\usepackage{multirow}
\usepackage{amssymb}
\usepackage{amsmath}
\usepackage{cite}
\usepackage{xcolor}
\begin{document}

\title{Graph Learning for Combinatorial Optimization: A Survey of State-of-the-Art 
}


\author{}

\author{Yun Peng         \and Byron Choi \and
        Jianliang Xu  
}


\institute{Yun Peng \at
              \email{yunpeng@comp.hkbu.edu.hk}           
           \and
           Byron Choi \at
              \email{bchoi@comp.hkbu.edu.hk}
           \and
           Jianliang Xu \at
              \email{xujl@comp.hkbu.edu.hk}
}

\date{Received: date / Accepted: date}

\maketitle

\newcommand{\yun}[1]{\textcolor{blue}{#1}}

\begin{abstract}
Graphs have been widely used to represent complex data in many applications, such as e-commerce, social networks, and bioinformatics. Efficient and effective analysis of graph data is important for graph-based applications. However, most graph analysis tasks are combinatorial optimization (CO) problems, which are NP-hard. Recent studies have focused a lot on the potential of using machine learning (ML) to solve graph-based CO problems. Most recent methods follow the two-stage framework. The first stage is graph representation learning, which embeds the graphs into low-dimension vectors. The second stage uses machine learning to solve the CO problems using the embeddings of the graphs learned in the first stage.
The works for the first stage can be classified into two categories, graph embedding methods and end-to-end learning methods. For graph embedding methods, the learning of the the embeddings of the graphs has its own objective, which may not rely on the CO problems to be solved. The CO problems are solved by independent downstream tasks. For end-to-end learning methods, the learning of the embeddings of the graphs does not have its own objective and is an intermediate step of the learning procedure of solving the CO problems.
The works for the second stage can also be classified into two categories, non-autoregressive methods and autoregressive methods. Non-autoregressive methods predict a solution for a CO problem in one shot. A non-autoregressive method predicts a matrix that denotes the probability of each node/edge being a part of a solution of the CO problem. The solution can be computed from the matrix using search heuristics such as beam search.  
Autoregressive methods iteratively extend a partial solution step by step. At each step, an autoregressive method predicts a node/edge conditioned to current partial solution, which is used to its extension. 
In this survey, we provide a thorough overview of recent studies of the graph learning-based CO methods.  The survey ends with several remarks on future research directions.
\end{abstract}

\newcommand{\tabincell}[2]{\begin{tabular}{@{}#1@{}}#2\end{tabular}}

\section{Introduction}

Graphs are ubiquitous and are used in a wide range of domains, from e-commerce \cite{GE4Rec}, \cite{GNNRec} to social networking~\cite{socialNetwork} \cite{hlx2019community} to bioinformatics \cite{GEbio}, \cite{Fingerprints}. Effectively and efficiently analyzing graph data is important for graph-based applications. However, many graph analysis tasks are combinatorial optimization (CO) problems, such as the traveling salesman problem (TSP) \cite{pointerNN}, maximum independent set (MIS) \cite{Stru2VRL}, maximum cut (MaxCut) \cite{maxcutNP}, minimum vertex cover (MVC) \cite{qifeng}, maximum clique (MC) \cite{MCsparse}, graph coloring (GC) \cite{VColor}, subgraph isomorphism (SI) \cite{subisoAuth}, and graph similarity (GSim) \cite{authSubSim}. These graph-based CO problems are NP-hard. In the existing literature on this subject, there are three main approaches used to solve a CO problem: exact algorithms, approximation algorithms, and heuristic algorithms. Given a CO problem on a graph $G$, exact algorithms aim to compute an optimum solution. Due to the NP-hardness of the problems, the worst-case time complexity of exact algorithms is exponential to the size of $G$. To reduce time complexity, approximation algorithms find a suboptimal solution that has a guaranteed approximation ratio to the optimum, with a worst-case polynomial runtime. Nevertheless, many graph-based CO problems, such as General TSP \cite{tspInapproximable}, GC \cite{GCInappro}, and MC \cite{MCInappro}, are inapproximable with such a bounded ratio. Thus, heuristic algorithms are designed to efficiently find a suboptimal solution with desirable empirical performance. Despite having no theoretical guarantee of optimality, heuristic algorithms often produce good enough solutions in practice.

The practice of applying machine learning (ML) to solve graph-based CO problems has a long history. For example, as far back as the 1980s, researchers were using the Hopfield neural network to solve TSP \cite{HopfieldNNTSP}, \cite{Smith99}. Recently, the success of deep learning methods has led to an increasing attention being paid to this subject \cite{MILP}, \cite{RLCO}, \cite{Stru2VRL}, \cite{pointerNN}. Compared to manual algorithm designs, 
 ML-based methods have several advantages in solving graph-based CO problems. First, ML-based methods can automatically identify distinct features from training data. In contrast, human algorithm designers need to study the heuristics with substantial problem-specific research based on intuitions and trial-and-errors. Second, for a graph-based CO problem, ML has the potential to find useful features that it may be hard to specify by human algorithm designers, enabling it to develop a better solution \cite{GCBetter}. Third, an ML-based method can adapt to a family of CO problems. For example, S2V-DQN \cite{Stru2VRL} can support TSP, MVC, and MaxCut; GNNTS \cite{qifeng} can support MIS, MVC, and MC. In comparison, it is unlikely for a handcrafted algorithm of one CO problem to be adapted to other CO problems.

Most recent graph learning-based CO methods follow the two-stage framework. The first stage is graph representation learning which embeds the graphs into low-dimension vectors. The second stage uses machine learning to solve the CO problems using the embedding vectors of the graphs learned in the first stage. In this survey, we review the state-of-the-art works of the two stages, respectively.

For the first stage, existing graph representation learning techniques that have been used in ML-based CO methods can be classified into two categories: graph embedding methods and end-to-end learning methods. On one hand, graph embedding methods embed the nodes of a graph into low-dimension vectors. The embedding vectors of the graph learned are inputted to downstream machine learning tasks to solve CO problems. Graph embedding has its learning objective, which may not rely on the CO problems to be solved. The embeddings of the graph are fixed during the solving of the downstream task. On the other hand, in end-to-end learning methods, graph representation learning does not have its own learning objective and is an intermediate step of the learning procedure of solving the CO problem. The embeddings learned are specific to the CO problem being solved.

For the second stage, existing works can be classified into two categories: non-autoregressive methods and autoregressive methods. On one hand, non-autoregressive methods predict a solution for a graph-based CO problem in one shot. For example, for the TSP problem on a graph, a non-autoregressive method predicts a matrix, where each element of the matrix is the probability of an edge belonging to a TSP tour. The TSP tour can be computed from the matrix using beam search.  On the other hand, autoregressive methods iteratively extend a partial solution step by step. For the TSP problem, at each step, an autoregressive method predicts an edge conditioned to the current partial TSP tour, which is used to extend the current partial TSP tour.

There have been several surveys of graph representation learning \cite{HamiltonSurvey}, \cite{CaiSurvey}, \cite{chenSurvey}, \cite{CuiPeng}, \cite{ZhangSurvey}. However, existing surveys mainly focus on the graph representation learning models and their applications in node classification, link prediction or graph classification. In contrast, we focus on using graph learning to solve CO problems.
There have also been several previous surveys that have discussed ML-based CO methods \cite{RLCO}, \cite{MILP}, \cite{LambNeuralSymbolic}. The present survey, however, has different emphases from previous studies. The survey \cite{MILP} focuses on branch and bound (B\&B) search techniques for the mixed-integer linear programming (MILP) problem. Although many graph-based CO problems can be formulated using MILP and solved using the B\&B method, most existing ML-based methods for solving graph-based CO problems focus on graph-specific methods. Mazyavkina et al. \cite{RLCO} discuss RL-based CO methods. However, there are ML-based CO methods that do not use RL. This survey is not limited to RL approaches. Lamb et al. \cite{LambNeuralSymbolic} survey the GNN-based neural-symbolic computing methods. Symbolic computing is a broad field and  graph-based CO is a topic of it. In contrast, we focus on the graph-based CO problems in this survey.

The rest of this survey is organized as follows. Section~\ref{back} presents the notations and preliminaries. Section~\ref{ga4gl} summarizes graph representation learning techniques. Section~\ref{gl4ga} discusses the use of ML to solve graph-based CO problems. Section~\ref{future} suggests directions for  future research. Section~\ref{conc} concludes this survey. Sec.~\ref{sec-abbr} lists the abbreviations used in this survey.

\section{Notations and Preliminaries}\label{back}

In this section, we present some of the notations and definitions that are  frequently used in this survey. 

We denote a graph by $G$ = $(V,E, {\bf X})$,
where $V$ and $E$ are the node set and the edge set of $G$,
respectively, ${\bf X}^{|V|\times d'}$ is the matrix of initial features of all nodes, and ${\bf x}_u$ = ${\bf X}$$[u,\cdot]$ denotes the initial features of node $u$. We may choose $u \in G$ or $u \in V$ to denote a node of the graph, when the choice is more intuitive. Similarly, we may use $(u,v) \in G$ or $(u,v) \in E$ to denote an edge of the graph. The adjacency matrix of $G$ is denoted by ${\bf A}$. The weight of edge $(u,v)$ is denoted by  
$w_{u,v}$. We use $\bf P$ to denote the transition matrix, where ${\bf P}[u,v] = w_{u,v}/\sum_{v'\in G} w_{u,v'}$. ${\bf P}^k = \prod_k {\bf P}$ is the $k$-step transition matrix and $\bf P$ is also called the 1-step transition matrix. We use $g$ to denote a subgraph of $G$ and $G\backslash g$ to denote the subgraph of $G$ after removing all nodes in $g$. For a node $u\in G,u\not\in g$, we use $g\cup\{u\}$ to denote adding the node $u$ and the edges $\{(u,v)|v\in g, (u,v)\in G\}$ to $g$.    
$G$ can be a directed or undirected graph. If $G$ is directed, $(u,v)$ and $(v,u)$ may not present simultaneously in $E$. $N^o(u)$ and $N^i(u)$  denote the outgoing and incoming neighbors of $u$, respectively. If $G$ is undirected, $N(u)$ denotes the neighbors of $u$. 
We use a bold uppercase character to denote a matrix ({\it e.g.}, $\bf X$), a bold lowercase character to denote a vector ({\it e.g.}, ${\bf x}$), and a lowercase character to denote a scalar ({\it e.g.}, $x$). The embedding vectors (or embeddings for short) of a node $u$ and a graph $G$ are $d$-dimensional vectors denoted by ${\bf h}_u$ and ${\bf h}_G$, respectively. 
Table~\ref{notation_table} summarizes the notations of frequently used symbols.

\begin{table}[t]
\vspace{0ex}
\caption{Notations of frequently used symbols and their meaning}
\centering
\begin{scriptsize}
\vspace{-0ex}
\begin{tabular}{|c|l|c|c|}
\hline
Notation & Description \\
\hline \hline
$G=(V,E,\bf{X})$ & a graph\\ \hline
$\bf A$ & the adjacent matrix of $G$\\ \hline
$u,v$ & node $u$ and node $v$ of $G$\\ \hline
${\bf X}$ & a matrix of features of all nodes\\ \hline
${\bf x}_u$ & the features of $u$, {\it i.e.}, a row of $\bf X$ \\\hline
${\bf y}$ & a signal, {\it i.e.}, a column of $\bf X$\\ \hline
$\mathcal N$$_u$ & neighborhood of $u$\\ \hline
${\bf h}_u, {\bf h}_G$ & embedding vector of a node $u$, a graph $G$\\ \hline
${\vec H}$ & the matrix of embedding vectors of all nodes \\ \hline
$d$  &   dimension of a vector \\ \hline
$f^l_\theta$ & the $l$-th layer of a neural network with parameter $\theta$\\ \hline
\end{tabular}\label{notation_table}
\end{scriptsize}
\vspace{0ex}
\end{table}

A graph-based CO problem is formulated in Definition~\ref{def:coproblem}.

\begin{definition}\label{def:coproblem}
Given a graph $G$ and a cost function $c$ of the subgraphs of $G$, a
CO problem is 
to find the optimum value of $c$ or the corresponding subgraph that produces that optimum value.
\end{definition}

For example, for a graph $G$, the maximum clique (MC) problem is to find the largest clique of $G$, and the minimum vertex cover (MVC) problem is to find the minimum set of nodes that are adjacent to all edges in $G$.

\subsection{Overview of Graph Learning Based CO Methods}

\begin{figure}
\centering
\includegraphics[width = 12cm]{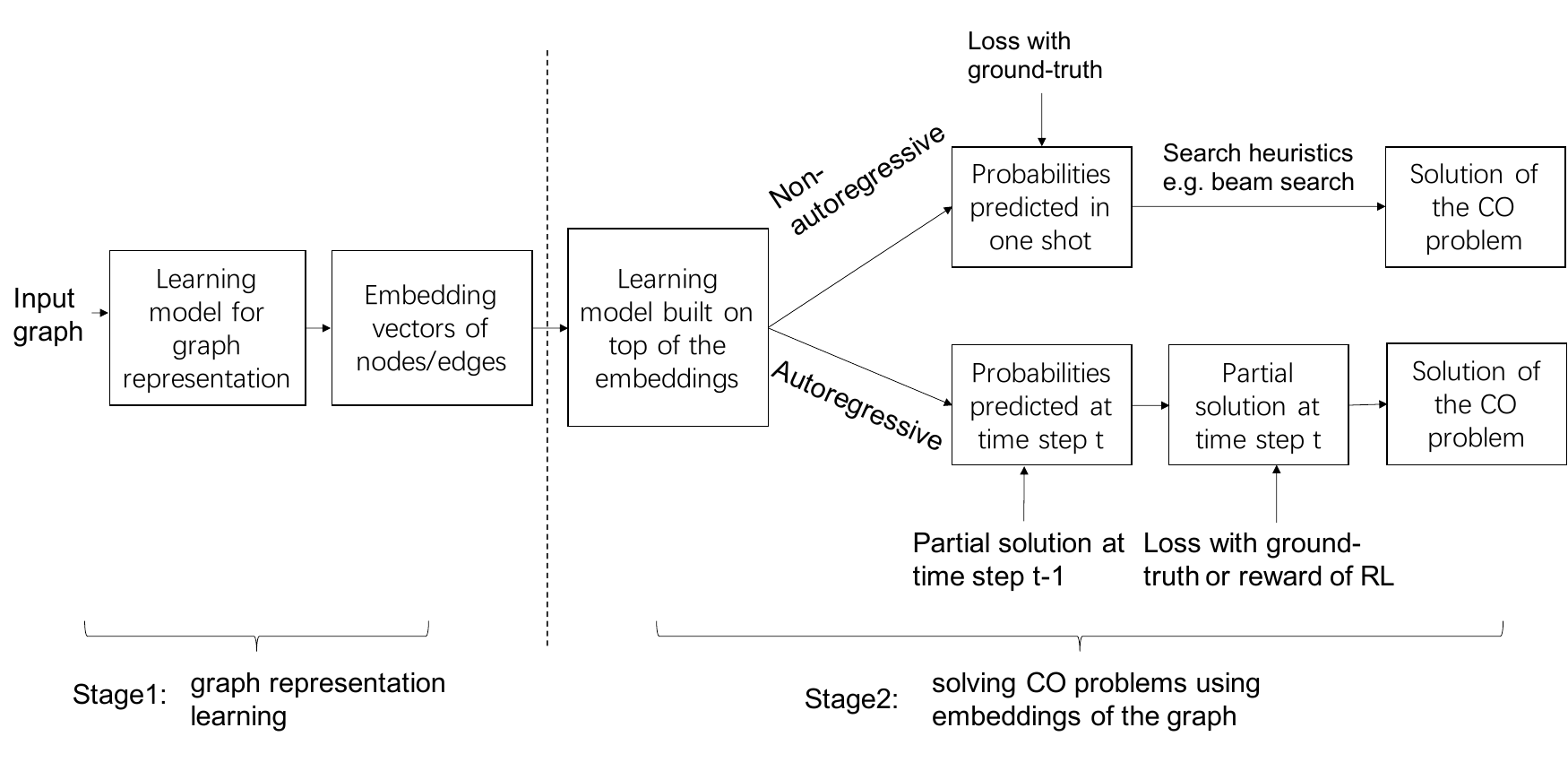}
\caption{Overview of the two stages of ML-based CO methods} 
\label{fig:overview}
\end{figure}

Most existing methods that use machine learning to solve the graph-based CO problem follow the two-stage framework, as illustrated in Fig.~\ref{fig:overview}. Given an input graph, the first stage is to learn the representation of the graph in a low-dimension embedding space. The nodes or edges of the graph are represented as embedding vectors (or {\it embeddings} for short). The techniques for the first stage are discussed in Sec.~\ref{ga4gl}. The second stage uses machine learning to solve the CO problem using the embeddings of the graph learned in the first stage. The techniques for the second stage are discussed in Sec.~\ref{gl4ga}.  

There are mainly two ways to learn graph representation in the first stage. In the first way, the embeddings are learned by graph embedding methods. Graph embedding has its own learning objectives that may not rely on the objective of the CO problem to be solved in the second stage. The CO problem is solved as a downstream task of graph embedding and the gradient of the loss of the CO problem in the second stage will not be back-propagated to the first stage. In the second way, the CO problem is solved in the end-to-end manner.  The first stage does not have its own learning objective and the gradient of the second stage is back-propagated to the first stage for learning the embeddings of the graph.

There are mainly two different approaches to solve the CO problems in the second stage, namely non-autoregressive methods and autoregressive methods.
The non-autoregressive methods predict a solution for a graph-based CO problem in one shot. A non-autoregressive method predicts a matrix that denotes the probability of each node/edge being a part of a solution. The solution of the CO problem can be computed from the matrix by 
search heuristics, {\it e.g.}, beam search. The autoregressive methods compute the solution by iteratively extending a partial solution. At each time step, the node/edge that is used to extend the partial solution is predicted conditioned to the current partial solution.

\section{Graph Representation Learning Methods} \label{ga4gl}

In this section, we survey the graph representation learning methods that have been applied to solve graph-based CO problems. In Sec.~\ref{grl:emb}, we review the graph embedding methods, which learn the embeddings of the graph independently to the downstream task of solving the CO problem; and in Sec.~\ref{grl:etoe}, we review the end-to-end learning methods that learn the embeddings of the graph as an intermediate step of solving the CO problem.

\subsection{Graph Embedding Methods}\label{grl:emb}

We first review generalized SkipGram and AutoEncoder that are two widely used models in graph embedding.

\subsubsection{Generalized SkipGram}

The generalized SkipGram model is extended from the well-known SkipGram model \cite{w2v}  for embedding words in natural language processing. The generalized SkipGram model relies on the neighborhood ${\mathcal N}_u$ of node $u$ to learn an embedding of $u$. The objective is to maximize the likelihood of the nodes in ${\mathcal N}_u$ conditioned on $u$.

\begin{equation}\label{loss1_skipgram}
\max {P}(v_1,v_2,...,v_{|{\mathcal N}_u|} | u), v_i\in {\mathcal N}_u
\end{equation}

Assuming conditional independence, ${P}(v_1,v_2,...,v_{|{\mathcal N}_u|} | u)$ = $\prod_{v_i\in {\mathcal N}_u} {P}(v_i | {\bf h}_u)$. $P(v_i | {\bf h}_u)$ can be defined as $\frac{{\bf h}_{v_i}^T {\bf h}_u}{\sum_{v\in G} {\bf h}_{v}^T {\bf h}_u}$.  Maximizing $\prod_{v_i\in {\mathcal N}_u} {P}(v_i | {\bf h}_u)$ is then equivalent to maximizing its logarithm. Hence, (\ref{loss1_skipgram}) becomes

\begin{equation}\label{loss2_skipgram}
\begin{array}{ll}
&  \max \sum_{v_i \in {\mathcal N}_u} \log {P}(v_i | {\bf h}_u)\\[0.2cm]
 & = \max \sum_{v_i \in {\mathcal N}_u} \log  \frac{{\bf h}_{v_i}^T {\bf h}_u}{\sum_{v\in G} {\bf h}_{v}^T {\bf h}_u}\\[0.2cm]
\end{array}
\end{equation}

Since computing the denominator of the softmax in (\ref{loss2_skipgram}) is time consuming, many optimization techniques have been proposed. Negative sampling \cite{w2v} is one of the most well-known techniques. Specifically, the nodes in the neighborhood ${\mathcal N}_u$ of $u$ are regarded as positive samples of $u$. On the other hand, the nodes not in ${\mathcal N}_u$ are considered negative samples of $u$. Then, maximizing the likelihood in Formula~\ref{loss2_skipgram} can be achieved as follows:

\begin{equation}\label{maxloss_skipgram}
\max \log\sigma({\bf h}_v^T {\bf h}_u) + \sum_{i=1}^K {\mathbb E}_{\bar{v}\sim { P_n}} \log\sigma (-{\bf h}_{\bar{v}}^T {\bf h}_u),
\end{equation}

\noindent
where $v$ is a positive sample of $u$, $\bar{v}$ is a negative sample, ${P_n}$ is the probability distribution of negative samples, $\bar{v}\sim {P_n}$ means sampling a node from the probability distribution $P_n$, $K$ is the number of negative samples, $\sigma$ is the sigmoid activation function, and $\mathbb E$ is the expectation.

To conveniently adopt the gradient descent algorithms, maximizing an objective is often rewritten as minimizing its negative. Thus, the objective function of the generalized SkipGram model is to minimize the loss $\cal L$ as follows:

\begin{equation}\label{loss3_skipgram}
\min {\mathcal L} = \min -\log\sigma({\bf h}_v^T {\bf h}_u) - \sum_{i=1}^K {\mathbb E}_{\bar{v}\sim { P_n}} \log\sigma (-{\bf h}_{\bar{v}}^T {\bf h}_u)
\end{equation}

Existing studies on the generalized SkipGram model define the neighborhood in different ways. For example, LINE \cite{LINE} defines the 1-hop neighbors as the neighborhood in order to preserve the second-order proximity; DeepWalk \cite{deepwalk} uses the random walk to define the neighborhood for preserving more global structural information of $G$.

\subsubsection{AutoEncoder}\label{GEMB_autoencoder}

AutoEncoder is composed of an encoder and a decoder. 
For a graph-based CO problem, the encoder encodes the nodes in the graph into $d$-dimensional embedding vectors. The decoder then predicts a solution to the CO problem using the node embeddings ({\it e.g.}, PointerNet \cite{pointerNN}).

Formally, the encoder is a function 
\[
enc: {\mathbb R}^{d'} \to {\mathbb R}^d
\]

\noindent
$enc({\bf x}_u)$ embeds node $u$  into ${\bf h}_u \in {\mathbb R}^{d}$. 

There are several different types of decoder. For instance, the inner product-based decoder, the reconstruction-based decoder, and the classification-based decoder are three widely-used decoders.

The inner product-based decoder is a function 
\[
dec: {\mathbb R}^d\times {\mathbb R}^d \to {\mathbb R}
\]

\noindent
$dec({\bf h}_u, {\bf h}_v)$ returns the similarity of ${\bf h}_u$ and ${\bf h}_v$. Let $sim(u,v)$ denote the proximity of $u$ and $v$ in $G$ ({\it e.g.}, $A[u,v]$ in \cite{SDNE}). The objective function of the inner product decoder is to minimize the loss 
\begin{equation}\label{equ:loss_inner_prod_ae}
{\mathcal L} = \sum_{(u,v)\in {\mathcal D}} dist(dec({\bf h}_u,{\bf h}_v), sim(u,v)),
\end{equation}

\noindent
where $\mathcal D$ is the training dataset and $dist$ is a user-specified distance function.

The reconstruction-based decoder is  a function 
\[
dec: {\mathbb R}^{d} \to {\mathbb R}^{d'}
\]

\noindent
$dec({\bf h}_u)$ outputs $\hat{{\bf x}}_u$ as the reconstruction of ${\bf x}_u$. The objective function is to minimize the reconstruction loss
\[
{\mathcal L} = \sum_{u\in G} ||(dec({\bf h}_u), {\bf x}_u)||^2_2
\]

The encoder and the decoder can be implemented by different types of neural networks, {\it e.g.}, the multi-layer perceptron (MLP) \cite{SDNE} or the recurrent neural network (RNN) \cite{DeepRecurNN}.

\subsubsection{Generalized SkipGram Based Graph Embedding Method}

This subsection reviews the generalized SkipGram based graph embedding methods DeepWalk \cite{deepwalk}, Node2Vec \cite{node2vec}, and Struc2Vec \cite{Struc2vec}; and the subgraph based graph embedding methods DeepGK \cite{dgk}, Subgraph2Vec \cite{subgraph2vec}, RUM \cite{RUM}, Motif2Vec \cite{motif2vec},  and MotifWalk \cite{MGCN} .

DeepWalk \cite{deepwalk} was one of the earlist works to introduce the generalized SkipGram model to graph embedding. The main idea of DeepWalk is to sample a set of truncated random walks of the graph $G$, and the nodes in a window of a random walk are regarded as co-occurence. The neighborhood of a node is the nodes that co-occurred with it. DeepWalk uses the generalized SkipGram model with the negative sampling (refer to Formula~\ref{loss3_skipgram}) to learn the graph embedding.

\begin{figure}
\centering
\includegraphics[width = 10cm]{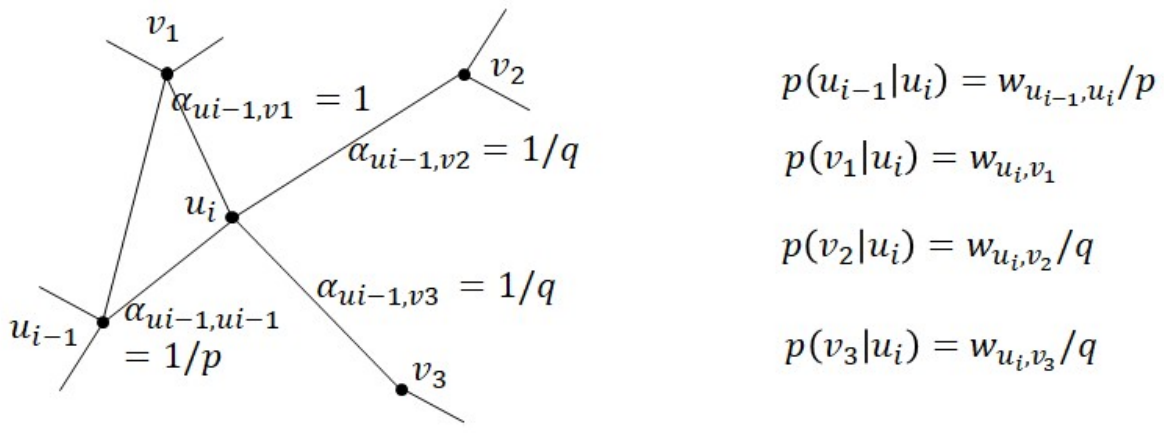}
\caption{An example of selecting the next node by the second-order random walk of Node2Vec \cite{node2vec}. $u_i$ is the current node of the random walk and $u_{i-1}$ is the previous node. $u_{i-1}$, $v_1$, $v_2$, and $v_3$ can be selected as the next node with the corresponding probabilities, respectively.}
\label{fig:node2vec}
\end{figure}

To incorporate more flexibility into the definition of node neighborhood, Node2Vec \cite{node2vec} introduces breadth-first search (BFS) and depth-first search (DFS) in neighborhood sampling. The nodes found by BFS and DFS can capture different structural properties. Node2Vec uses the second-order random walk to simulate the BFS and DFS. (``second-order'' means that when the random walk is at the step $i$, the random walk needs to look back to the step $i-1$ to decide the step $i+1$.) Two parameters $p$ and $q$ are introduced to control the random walk. $p$ controls the probability of return to an already visited node in the following two steps; and $q$ controls the probability of visiting a close or a far node in the following two steps. Let $u_{i}$ denote the current node in the walk and $u_{i-1}$ denote the previous node. The probability of the random walk to visit the next node $u_{i+1}$ is defined as below.

\begin{equation}\label{def:n2v2ndrw}
{P}(u_{i+1}|u_{i}) = 
\begin{cases}
\alpha_{u_{i-1},u_{i+1}}\times w_{u_i,u_{i+1}} & \text{if } (u_{i},u_{i+1})\in G\\
0 & \text{otherwise}
\end{cases}
\end{equation}

\[
\alpha_{u_{i-1},u_{i+1}} = 
\begin{cases}
1/p & \text{if } dist(u_{i-1},u_{i+1}) = 0\\
1 & \text{if } dist(u_{i-1},u_{i+1}) = 1\\
1/q & \text{if } dist(u_{i-1},u_{i+1}) = 2\\
\end{cases}
\]
where $dist(u_{i-1},u_{i+1})$ is the shortest distance from $u_{i-1}$ to $u_{i+1}$ and $w_{i,i+1}$ is the weight of the edge $(u_{i},u_{i+1})$.
An example is shown in Fig.~\ref{fig:node2vec}. The current node of the random walk is $u_i$. There are four nodes $u_{i-1}$, $v_1$, $v_2$ and $v_3$ that can be the next node of the random walk. The probability of selecting each of them as the next node is shown in Fig.~\ref{fig:node2vec}.  

Struc2Vec \cite{Struc2vec} argues that the random walks of Node2Vec cannot find nodes that have similar structures but are far away. Struc2Vec builds a multi-layer graph $G'$ for the input graph $G$. The layer $l$ is a complete graph $G'_l$, where each node in $G$ is a node in $G'_l$ and each edge $(u,v)\in G'_l$ is weighted by the structural similarity of the $l$-hop neighborhoods of $u$ and $v$ in $G$. In this way, two nodes that are far away in $G$ can reach each other by just one hop in $G'_l$. The nodes in $G'_l$ can have directed edges to the nodes in $G'_{l-1}$ and $G'_{l+1}$. Random walks are sampled on $G'$, and the generalized SkipGram model is used to learn the node embedding. 

Besides using paths to sample the neighborhood, many works use representative subgraphs of the input graph. The representative subgraphs may be termed motifs, graphlets or kernels in different studies. Yanardag and Vishwanathan \cite{dgk} propose DeepGK, which is the earlist work embedding the motifs. The neighborhood of a motif $g$ is defined as the motifs within a small distance from $g$. The generalized SkipGram model is used to learn the embeddings for the motifs. 

Yu et al. \cite{RUM} propose a network representation learning method using motifs
(RUM). RUM builds a motif graph $G'$ for the input graph $G$, where each node in $G'$ is a motif of $G$ and two nodes have an edge in $G'$ if the corresponding motifs share common nodes. Triangle is used as the graph motif in RUM. RUM uses random walks on the motif graph $G'$ to define the neighborhood of a motif. Then, the generalized SkipGram model is used to learn the embedding of the motif. An original node $u$ of $G$ may occur in multiple motifs of $G'$. RUM uses the average of the embeddings of the motifs as the embedding of $u$.

Dareddy et al. \cite{motif2vec} propose another type of motif graph. Given a graph $G=(V,E)$, for each motif $g$, Motif2Vec builds a motif graph $G'=(V,E')$, where the weight of an edge $(u,v)\in E'$ is the number of motif instances of $g$ in $G$ that contain node $u$ and $v$. Then, Motif2Vec simulates a set of random walks on each motif graph and uses Node2Vec \cite{node2vec} to learn the embeddings of the nodes in $G$.
A similar idea is also proposed in the MotifWalk method of \cite{MGCN}.

Narayanany et al. \cite{subgraph2vec} propose Subgraph2Vec to compute the embeddings of the neighboring subgraphs of the nodes in the input graph. Let $g_u$ denote the neighboring subgraph of a node $u$, Subgraph2Vec computes ${\bf h}_{g_u}$ using the generalized SkipGram model. The neighborhood of $g_u$ is defined as the neighboring subgraphs of the neighbors of $u$, {\it i.e.}, $\{g_v| v\in N(u)\}$.

\subsubsection{AutoEncoder Based Graph Embedding}

AutoEncoder-based graph embedding often preserves the graph structure properties measured by the following proximities.

\begin{definition}
Given a graph $G=(V,E)$, the {\it first-order proximity} from $u$ to $v$ is the weight of $(u,v)$. If $(u,v)\in E$, $p^{(1)}(u,v)=w_{u,v}$; otherwise, $p^{(1)}(u,v)=0$.
\end{definition}

The first-order proximity captures the direct relationship between nodes. The second-order proximity captures the similarity of the neighbors of two nodes.

\begin{definition}\label{def:2ndprox}
Given a graph $G$, the {\it second-order proximity} between $u$ and $v$ is $p^{(2)}(u,v)$ = $sim({\bf p}^{(1)}(u), {\bf p}^{(1)}(v))$, where ${\bf p}^{(1)}(u)$ is the vector of the first-order proximity from $u$ to all other nodes in $G$, i.e., ${\bf p}^{(1)}(u)=(p^{(1)}(u,v_1), p^{(1)}(u,v_2), ...,$ $p^{(1)}(u,v_{|V|}))$, and $sim$ is a user-specified similarity function.
\end{definition}

The first- and second-order proximities encode the local structures of a graph. Proximities to capture more global structures of a graph have also been proposed in the literature. For example, Cai et al. \cite{CaiSurvey} propose to use $p^{(k)}(u,v)$ (recursively defined, similar to Definition~\ref{def:2ndprox}) as the $k$-th-order proximity between $u$ and $v$, Cao et al. \cite{GraRep} use the $k$-step transition probability $\bf P$$^k[u,v]$ to measure the $k$-step relationship from $u$ to $v$, Chen et al. \cite{GraphCSC} use the node centrality, Tsitsulin et al. \cite{VERSE} use the Personalized PageRank, and Ou et al. \cite{HOPE} use the Katz Index and Adamic-Adar to measure more global structural properties of $G$.

Large-scale information
network embedding (LINE) \cite{LINE} preserves both the first- and second-order proximity in graph embedding using two AutoEncoders, respectively. In order for AutoEncoder to preserve the first-order proximity, the encoder is a simple embedding lookup \cite{chamiSurvey}. The decoder outputs the estimated adjacent matrix using the node embeddings, and the objective is to minimize the loss between the estimated adjacent matrix and the ground truth.

The decoder of LINE is designed as follows. Since adjacent nodes $u$ and $v$ in $G$ have high first-order proximity, they should be close in the embedding space. LINE uses the inner product of ${\bf h}_u$ and ${\bf h}_v$ to measure the distance between $u$ and $v$ in the embedding space, as shown below.

\begin{equation}
{P}_1(u,v) = \frac{1}{1+exp(-{\bf h}_u^T {\bf h}_v)}
\end{equation}

$P_1(\cdot,\cdot)$ defines the estimated distribution of the first-order proximity ({\it i.e.}, the estimated adjacent matrix). LINE ensures that the estimated distribution $P_1(\cdot,\cdot)$ is close to the empirical distribution $\hat{P}_1(\cdot,\cdot)$ so as to preserve the first-order proximity. 

\begin{equation}\label{edgeprob}
{\mathcal L}_1 = \min dist(\hat{P}_1(\cdot,\cdot), {P}_1(\cdot,\cdot))
\end{equation}

\noindent
where $\hat{P}_1(u,v)=\frac{w_{u,v}}{\sum_{(u',v')\in G} w_{u',v'}}$ and $dist$ is the distance between two probability distributions. If the KL-divergence is used as $dist$, ${\mathcal L}_1$ becomes

\begin{equation}\label{firstproxloss}
{\mathcal L}_1 = \min -\sum_{(u,v)\in G} w_{u,v} \log {P}_1(u, v)
\end{equation}

In order for AutoEncoder to preserve the second-order proximity, the encoder is a simple embedding lookup \cite{chamiSurvey}. The decoder outputs an estimated distribution between each node and its neighbors. The estimated distribution is reconstructed from the embeddings of the nodes. The objective is to minimize the reconstruction loss between the estimated distribution and the ground truth.

The decoder is designed as follows.
Inspired by word embedding \cite{w2c}, the neighbors of $u$ are regarded as the ``context'' of $u$. LINE uses a conditional probability $P_2(v|u)$ defined in Formula~\ref{neighprob}  to model the estimated probability of $u$ generating a neighbor $v$. 

\begin{equation}\label{neighprob}
{P}_2(v | u) = \frac{exp({\bf h}_v'^T {\bf h}_u)}{\sum_{v'\in G} exp({\bf h}_{v'}'^T {\bf h}_u)},
\end{equation}

\noindent
where ${\bf h}'$ is the vector of a node when the node is regarded as context. 

${P}_2(\cdot | u)$ defines the estimated distribution of $u$ over the context. The nodes $u$ and $u'$ in $G$ that have a high second-order proximity should have similar estimated distributions over the context, {\it i.e.}, ${P}_2(\cdot | u)$ should be similar to ${P}_2(\cdot | u')$. This can be achieved by minimizing the distance between the estimated distribution ${P}_2(\cdot | u)$ and the empirical distribution $\hat{P}_2(\cdot |u)$,  for each node $u$ in $G$. The empirical distribution $\hat{P}_2(\cdot|u)$ is defined as $\hat{P}_2(v|u) = w_{u,v} / \sum_{u,v'} w_{u,v'}$. LINE preserves the second-order proximity as follows. 

\begin{equation}\label{secondproxloss_general}
{\mathcal L}_2  = \min \sum_{u\in G} dist(\hat{P}_2(\cdot|u), {P}_2(\cdot | u)))
\end{equation}

Using the KL-divergence for $dist$, Formula~\ref{secondproxloss_general} produces

\begin{equation}\label{secondproxloss}
{\mathcal L}_2 = \min -\sum_{(u,v)\in G} w_{u,v} \log {P}_2(v|u)
\end{equation}

LINE trains the two AutoEncoders separately. The node embeddings generated by the two AutoEncoders are concatenated as the embeddings of the nodes. The model of LINE is also adopted by Tang et al. \cite{PTE} to embed the words in a heterogeneous text graph.

Wang et al. \cite{SDNE} argue that LINE is a shallow model, in the sense that it cannot effectively capture the highly non-linear structure of a graph. Therefore, structural deep network embedding (SDNE) is proposed as a mean of using the deep neural network to embed the nodes. As with LINE, SDNE also preserves the first- and second-order proximity. Both the encoder and decoder of SDNE are MLPs. Given a graph $G$, the encoder embeds ${\bf x}_u$ to ${\bf h}_u$, where ${\bf x}_u$ is the $u$-th row in the adjacent matrix $\vec A$ of $G$, and the decoder reconstructs $\hat{{\bf x}}_u$ from ${\bf h}_u$. 

SDNE preserves the first-order proximity by minimizing the distance in the embeded space for the adjacent nodes in $G$. 

\[
{\mathcal L}_1 = \sum_{(u,v)\in G} {\bf A}[u,v]\times ||{\bf h}_u - {\bf h}_v||^2_2
\]

The second-order proximity is preserved by minimizing the reconstrucspation loss.

\[
{\mathcal L}_2 = \sum_{u\in G} ||\hat{{\bf x}}_u - {\bf x}_u||_2^2
\]

SDNE combines ${\mathcal L}_1$, ${\mathcal L}_2$, and a regularizer term as the objective function and jointly optimizes them by means of a deep neural network. The first- and second-order proximity are preserved and the graph embedding learned is more robust than LINE. As demonstrated in experiments, SDNE outperforms LINE in several downstream tasks ({\it e.g.}, node classification and link prediction).

Versatile graph embedding method (VERSE) \cite{VERSE} shows that the first- and second-order proximity are not sufficient to capture the diverse forms of similarity relationships among nodes in a graph.
Tsitsulin et al. \cite{VERSE} propose to use a function $sim(u,v)$ to measure the similarity between any two nodes $u$ and $v$ in $G$, where $sim(\cdot, \cdot)$ can be any similarity function. The similarity distribution of $u$ to all other nodes can be defined by $sim(u,\cdot)$. The encoder of VERSE is a simple embedding lookup. The decoder estimates the similarity distribution using the node embeddings, as in Formula~\ref{neighprob}. The objective is to minimize the reconstruction loss  between the estimated similarity distribution  and the ground truth.

Dave et al. \cite{dave2019neural} propose Neural-Brane to capture both node attribute information and graph structural information in the embedding of the graph. Bonner et al. \cite{bonner2019exploring} study the interpretability of graph embedding models.

\subsubsection{Discussion}

The generalized SkipGram model is inspired by the word embedding model in natural language processing (NLP). Random walks of the graphs, which are the analog of sentences in texts are widely used by the generalized SkipGram model-based methods for computing the embeddings of the graphs. However, computing random walks is time-consuming. Moreover, the generalized SkipGram model is often regarded as a shallow model when compared to AutoEncoder. AutoEncoder can be deeper by stacking more layers and has more potentials to encode the complex and non-linear relationships between the nodes of a graph \cite{SDNE}. Recent works of word embedding in NLP also verify the advantage of AutoEncoder \cite{devlin-etal-2019-bert}. However, designing the architectures of the encoder and decoder and the loss function to encode the structure information of the graph is challenging.

Graph embedding methods can precompute the embedding vectors of graphs. The advantage is that the structure information encoded in the embeddings can be transferred to different downstream tasks. Graph embedding methods learn the embeddings of the graph without considering the downstream CO problems to be solved. The embeddings may not encode the information that are critical for solving the CO problem. There is an opportunity that the performance of the graph embedding-based methods may be inferior to the end-to-end learning methods for solving CO problems. Therefore, there have been recent studies on alternative graph representation learning methods for solving CO problems such as end-to-end learning methods.

\subsection{End-to-end Method}\label{grl:etoe}

Graph neural network (GNN) and AutoEncoder are widely used in the end-to-end learning methods of solving CO problems, where  computing the embeddings of the graphs is an intermediate step. 

\subsubsection{Graph Neural Network}

Graph neural network uses the graph convolution operation to aggregate graph structure and node content information. Graph convolution can be divided into two categories: i) spectral convolutions, defined using the spectra of a graph, which can be computed from the eigendecomposition of the graph's Laplacian matrix, and ii) spatial convolutions, directly defined on a graph by information propagation. 

\vspace{1ex}
\noindent
{\it A) Graph Spectral Convolution}
\vspace{1ex}

Given an undirected graph $G$, $\bf L$ = $\bf I$ - ${\bf D}^{-1/2} {\bf A D}^{-1/2}$ is the normalized Laplacian matrix of $G$. $\bf L$ can be decomposed into $\bf L$ = $\bf U\Lambda U$$^T$, where $\bf U$ is the eigenvectors ordered by eigenvalues, $\bf \Lambda$ is the diagonal matrix of eigenvalues, and $\bf \Lambda$$[i,i]$ is the $i$-th eigenvalue $\lambda_i$.

The {\it graph convolution} $\ast_G$ of an input signal $\bf s$ $\in {\mathbb R}^{|V|}$ with a filter ${\bf g_\theta}$ is defined as 

\begin{equation}\label{def:gspectral}
{\bf s} \ast_G {\bf g_\theta} = {\bf U} {\bf g_\theta} {\bf U}^T {\bf s}
\end{equation}

Existing studies on graph spectral convolution all follow Formula~(\ref{def:gspectral}), and the differences are the choice of the filter ${\bf g_\theta}$ \cite{WuSurvey}. The $u$-th row of the output channel is the embedding ${\bf h}_u$ of a node $u$.

\vspace{1ex}
\noindent
{\it B) Graph Spatial Convolution}
\vspace{1ex}

 Graph spatial convolution aggregates the information from a node's local neighborhood. Intuitively, each node sends messages based on its current embedding and updates its embedding based on the messages received from its local neighborhood. A graph spatial convolution model often stacks multiple layers, and each layer performs one iteration of message propagation. To illustrate this, we recall the definition given in GraphSAGE \cite{GraphSAGE}. A layer of GraphSAGE is as follows:

\begin{equation}\label{gconv}
{\bf h}_u^l = \sigma({\bf W}^l [{\bf h}_u^{l-1} || {\bf h}_{{\mathcal N}_u}^{l}])
\end{equation}

\begin{equation}
{\bf h}_{{\mathcal N}_u}^{l} = AGG( \{ {\bf h}_{v}^{l-1}, v\in {\mathcal N}_u \} ),
\end{equation}

\noindent
where 
$l$ denotes the $l$-th layer, $||$ denotes concatenation, ${\mathcal N}_u$ is a set of randomly selected neighbors of $u$, and $AGG$ denotes an order-invariant aggregation function. GraphSAGE suggests three aggregation functions: element-wise mean, LSTM-based aggregator, and max-pooling.

\subsubsection{GNN Based Graph Representation Learning}

Graph convolutional network (GCN) \cite{GCN} is a well-known graph spectral convolution model, which is an approximation of the original graph spectral convolution defined in Formula~\ref{def:gspectral}. Given a graph $G$ and a one-channel input signal $\vec s$ $\in {\mathbb R}^{|V|}$, GCN can output a $d$-channel signal ${\bf H}^{|V|\times d}$ as follows: 

\begin{equation}\label{equ:gcn_oneChannel}
 {\bf H} = {\bf s} \ast_G {\bf g_\theta} =  ({\bf \widetilde{D}}^{-1/2}{\bf \widetilde{A}}{\bf \widetilde{D}}^{-1/2}){\bf s} {\vec \theta},
\end{equation}


\noindent
where ${\vec \theta}$ is a ${1\times d}$  trainable parameter vector of the filter, ${\bf \widetilde{A}} = {\bf A} + {\bf I}$ and ${\bf \widetilde{D}}$ is a diagonal matrix with ${\bf \widetilde{D}}[i,i]$ = $\sum_j {\bf \widetilde{A}}[i,j]$. The $u$-th row of $\bf H$ is the embedding of the node $u$, ${\bf h}_u$.  To allow a $d'$-channel input signal $\bf S$$^{|V|\times d'}$ and output a $d$-channel signal ${\vec H}^{|V|\times d}$, the filter needs to take a parameter matrix ${\bf \Theta}^{d'\times d}$. Formula~\ref{equ:gcn_oneChannel} becomes

\begin{equation}\label{equ:mchannelGCN}
{\bf H} = {\bf S} \ast_G {\bf g_\Theta} = ({\bf \widetilde{D}}^{-1/2}{\bf \widetilde{A}}{\bf \widetilde{D}}^{-1/2}){\bf S} {\bf \Theta}.
\end{equation}

Let ${\bf s}_i$ denote the $i$-th channel ({\it i.e.}, column) of $\vec S$. ${\bf h}_u$ can then be written in the following way. 

\begin{equation}\label{equ:gcn}
{\bf h}_u = {\bf \Theta}^T {\bf y}, {\bf y}[i] = \sum_{v\in N_u\cup\{u\}} \frac{1}{\sqrt{|N_u|}\sqrt{|N_v|}} {\bf s}_i[v], 1\leq i\leq d',
\end{equation}

\noindent
where $\vec y$ is a $d'$-dimensional column vector.

When multi-layer models are considered, Formulas~\ref{equ:mchannelGCN} and \ref{equ:gcn} are written as Formulas~\ref{equ:mchannelGCN_layerwise} and \ref{equ:gcn_layerwise}, respectively, where $l$ denotes the $l$-th layer.

\begin{equation}\label{equ:mchannelGCN_layerwise}
{\bf H}^l = ({\bf \widetilde{D}}^{-1/2}{\bf \widetilde{A}}{\bf \widetilde{D}}^{-1/2}){\bf H}^{l-1} {\bf \Theta}^{l}
\end{equation}

\begin{equation}\label{equ:gcn_layerwise}
{{\bf h}^{l}}_{u} = {{\bf \Theta}^{l}}{}^{T} {\bf y}^{l}
\end{equation}


\[
\begin{array}{ll}
{\bf y}^l[i] & = \sum_{v\in N_u\cup\{u\}} \frac{1}{\sqrt{|N_u|}\sqrt{|N_v|}} {\bf H}^{l-1}[v,i]\\[0.4cm]
& = \sum_{v\in N_u\cup\{u\}} \frac{1}{\sqrt{|N_u|}\sqrt{|N_v|}} {\bf h}_v^{l-1}[i]
\end{array}
\]

From Formula~\ref{equ:gcn_layerwise}, we can observe that GCN  aggregates weighted information from a node's neighbors. In particular, for a node $u$ and a neighbor $v$ of $u$, the information from $v$ is weighted by their degrees, {\it i.e.}, $1/\sqrt{|N_u||N_v|}$. Graph attention network (GAT) \cite{GAT} argues that the fixed weight approach of GCN may not always be optimal. Therefore, GAT introduces the attention mechanism to graph convolution. A learnable weight function $\alpha(\cdot, \cdot)$ is proposed, where $\alpha(u,v)$ denotes the attention weight of $u$ over its neighbor $v$. Specifically, the convolution layer of GAT is as follows.

\begin{equation}\label{equ:attention}
{\bf h}_{u}^{l} = \sigma( \sum_{v\in N(u)} \alpha^l(u, v) {\bf W}^l {\bf h}_{v}^{l-1} )
\end{equation}

\begin{equation}
\alpha^l(u, v) =  \frac{exp( LeakyReLU({{\bf a}^l}{}^T[{\bf W}^l {\bf h}_{u}^{l-1} || {\bf W}^l {{\bf h}_v}^{l-1}]))}{\sum_{v'\in N(u)} exp( LeakyReLU({{\bf a}^l}{}^T [{\bf W}^l {\bf h}_{u}{}^{l-1} || {\bf W}^l {\bf h}_{v'}{}^{l-1}]) )},
\end{equation}

\noindent
where $||$ denotes concatenation, ${\bf a}^l$ and ${\bf W}^l$ are the trainable vector and matrix of parameters, respectively. 

The attention mechanism enhances models' capacity, and hence, GAT can perform better than GCN in some downstream tasks ({\it e.g.}, node classification). However, when $L$ layers are stacked, the $L$-hop neighbors of a node are needed to be computed.  If the graph $G$ is dense or a power-law graph, there may exist some nodes that can access almost all nodes in $G$, even for a small value of $L$. The time cost can be unaffordable.

\begin{figure}
\centering
\includegraphics[width = 10cm]{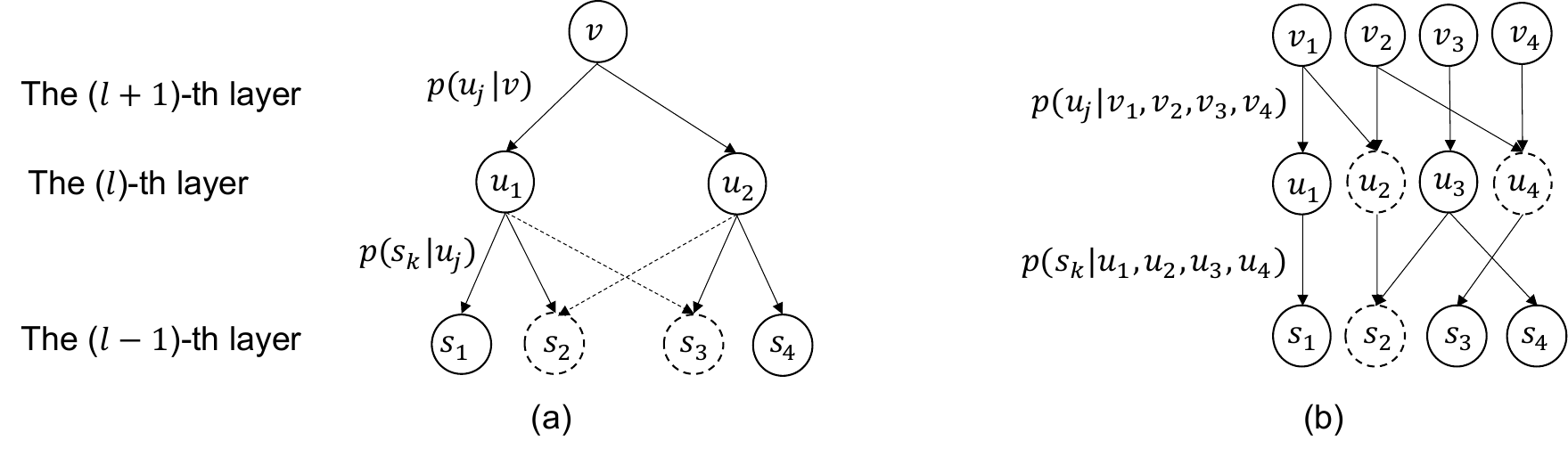}
\caption{Adaptive sampling of ASGCN \cite{ASGCN}: (a) the node-wise sampling and (b) 
the layer-wise sampling. In the node-wise sampling, each node in a layer samples its neighbors in the next layer independently. In particular, a node $v$ in the $l+1$-th layer samples its neighbors in the $l$-th layer by $p(u_j|v)$. In contrast, all nodes in a layer jointly sample the neighbors in the next layer. $u_j$ is sampled based on $p(u_j|v_1,v_2,...,v_4)$. The layer-wise sampling is more efficient than the node-wise sampling.} 
\label{fig:asgcn}
\end{figure}

To optimize efficiency, Hamilton et al. \cite{GraphSAGE} propose a sampling based method (GraphSAGE). GraphSAGE  randomly samples $k$ neighbors in each layer. Therefore, a model having $L$ layers only needs to expand $O(k^L)$ neighbors. Huang et al. \cite{ASGCN} further improve the sampling process with an adaptive sampling method. The adaptive sampling in \cite{ASGCN} samples the neighbors based on the embedding of $u$, as illustrated in Fig.~\ref{fig:asgcn}(a). The efficiency is further improved by layer-wise sampling, as shown in Fig.~\ref{fig:asgcn}(b). These sampling techniques are experimentally verified effective regarding the classification accuracy. 

\begin{figure}
\centering
\includegraphics[width = 12cm]{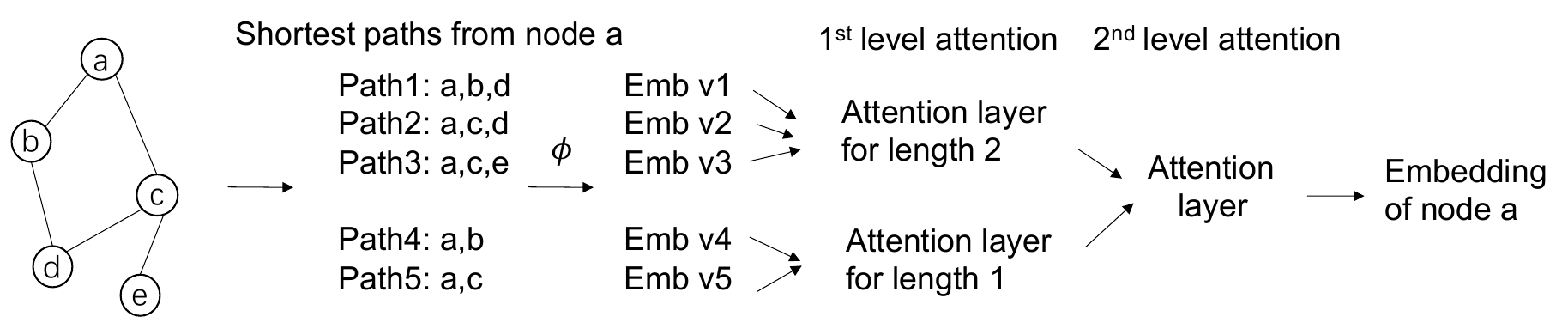}
\caption{The two-level convolution of SPAGAN \cite{SPAGAN}}
\label{fig:spagan}
\end{figure}

Yang et al. \cite{SPAGAN} combine the ideas of attention and sampling and propose the shortest path attention method (SPAGAN). The shortest path attention of SPAGAN has two levels, as shown in Fig.~\ref{fig:spagan}. The first level is length-specific, which embeds the shortest paths of the same length $c$ to a vector ${\bf h}_u^c$. The second level aggregates ${\bf h}_u^c$ of different values of $c$ to get the embedding ${\bf h}_u$ of $u$.

More specifically, let $P_u^c$ be the set of shortest paths starting from $u$ of the length $c$ and $p_{u,v}$ be a shortest path from node $u$ to node $v$. ${\bf h}_{u}^c$ is computed as follows.

\[
{\bf h}_{u}^c = \sum_{p_{u,v}\in P_u^c} \alpha_{u,v}\phi(p_{u,v}),
\]
where $\alpha_{u,v}$ is the attention weight and $\phi(p_{u,v})$ is a mean pooling that computes the average of the embeddings of the nodes in $p_{u,v}$.

\[
\alpha_{u,v} = \frac{exp(\sigma({\bf a}_1 [ ({\bf W} {\bf h}_u)||\phi(p_{u,v})])}{\sum_{p_{u,v'}\in P_u^c} exp(\sigma({\bf a}_1 [ ({\bf W} {\bf h}_u)||\phi(p_{u,v'})])},
\]
where $\vec a_1$ and $\bf W$ are trainable parameters shared by all nodes, and $||$ is concatenation.
The second level aggregates the paths with different lengths as follows.

\[
{\bf h}_{u} = \sigma(\sum_{c=2}^C \beta_c {\bf h}_{u}^c),
\]
where $C$ is a hyperparameter of the path length limit and $\beta_c$ is the attention weight.

\[
\beta_{c} = \frac{exp(\sigma({\bf a}_2 [ ({\bf W} {\bf h}_{u})||{\bf h}_{u}^c]))}{\sum_{c'=2}^C exp(\sigma({\bf a}_2 [ ({\bf W} {\bf h}_{u})||{\bf h}_{u}^{c'}]))},
\]
where ${\bf a}_2$ is a trainable parameter vector.

\subsubsection{AutoEncoder Based Graph Representation Learning}\label{end2end_autoencoder}

For the AutoEncoder used in the end-to-end learning, the embeddsings of the graph are computed by the encoder.  The decoder outputs the probabilities of nodes/edges belonging to the solutions of the CO problems. In recent works, the encoder mainly uses RNN and attention-based model, and the decoder mainly uses MLP, RNN and attention-based model. The encoder corresponds to the first stage of the ML-based CO methods and the decoder corresponds to the second stage (see Fig.~\ref{fig:overview}). In this subsection, we mainly focus on the encoder. The details of the decoders will be discussed in Section~\ref{gl4ga}.

The pointer network (Ptr-Net) proposed by Vinyals et al. \cite{pointerNN} is a seminal work of using AutoEncoder to solve the TSP problem. The encoder of Ptr-Net is an RNN taking the nodes of the graph $G$ as input and outputting an embedding of $G$, where the order of the nodes is randomly chosen. Experiments of Ptr-Net observe that the order of input nodes have affects on the quality of the TSP tour found. Therefore, the decoder of Ptr-Net introduces an attention mechanism that can assign weights to the input nodes and ignore the order of them.

Kool et al. \cite{kool2018attention} use AutoEncoder to sequentially output a TSP tour of a graph $G$. 
The encoder stacks $L$ self-attention layers. Each layer is defined as follows. 

\[
\hat{{\bf h}}_{v_i} = BN^l({\bf h}_{v_i}^{l-1}+MHA_i^l({\bf h}_{v_1}^{l-1},...,{\bf h}_{v_n}^{l-1}))
\] 

\[
{\bf h}_{v_i}^l = BN^l(\hat{{\bf h}}_{v_i}+MLP^l(\hat{{\bf h}}_{v_i})),
\]

\noindent
where ${\bf h}_i$ denotes the embedding vector of the node $v_i$, $l$ means the $l$-th layer, $MHA$ denotes the multi-head attention and $BN$ denotes the batch normalization. The embedding of $G$ ${\bf h}_G = \frac{1}{n} \sum_i^n {\bf h}_{v_i}^L$ and the embedding of each node ${\bf h}_{v_i}^L$ are input to the decoder.

\subsubsection{Discussions}

Most graph neural network-based methods adopt the message propagation framework. Each node iteratively aggregates the message from neighbors. The structure information of $k$-hops of a node can be captured by $k$ iterations of message aggregation. GNN does not require any node order and can support permutation invariance of CO problems. AutoEncoder-based methods are often used in solving the CO problems having sequential characteristics, {\it e.g.}, TSP. Sequence model is often used as the encoder to compute the embeddings of the graphs. Attention mechanism is used to support permutation invariance.

End-to-end learning methods learn the embeddings of graph as an intermediate step in solving the CO problem. The embeddings of the graph learned are more specific for the CO problem being solved and are expected to lead to better solutions of the CO problem.

A disadvantage of the GNN-based method is that the GNN is often shallow, due to the over-smooth problem. The attention-based encoder can alleviate this problem, where the encoder with self-attention layers and skip connections can be potentially deeper. However, the time complexity of such encoder on large graphs will be a bottleneck.

For the GNN-based method, the current trend is to use anisotropy GNN ({\it e.g.} GAT \cite{GAT}), which can differentiate the information propagated from different neighbors. For AutoEncoder-based method, more recent studies are integrating the attention mechanism with the sequence model to increase the capacity of the model and encode inductive biases.

\section{Graph Learning Based Combinatorial Optimization Methods} \label{gl4ga}

In this section, we review the works that solve CO problems using graph learning. We review the whole learning procedure in solving a CO problem. For the two stages of the learning procedure, we  pay more attention to the second stage, as the first stage has been thoroughly reviewed in the previous section. We will brief the first stage of the ML-based CO methods for the convenience of presentation.

Recent works can be classified into two categories. The first category is the non-autoregressive method which predicts the solution of a CO problem in one shot. The non-autoregressive method predicts a matrix that denotes the probability of each node/edge being a part of a solution. The solution of the CO problem can be found by search heuristics such as beam search. The second category is the autoregressive method, which constructs a solution by iteratively extending a partial solution to obtain a solution of the CO problem. Table~\ref{gpdlcompare} lists the selected graph learning-based CO methods.

Sec.~\ref{nonauto} summarizes the recent non-autoregressive methods for traver travelling salesman problem (TSP), graph partition, graph similarity, minimum vertex cover (MVC), graph coloring, maximum independent set, graph matching and graph isomorphism. Sec.~\ref{autoreg} presents the recent autoregressive methods for TSP, graph matching, graph alignment, MVC and maximum common subgraph.

\begin{table}[t]
\vspace{0ex}
\caption{Summary of selected CO methods using graph embedding}
\centering
\begin{scriptsize}
\vspace{-0ex}
\begin{tabular}{|c|c|c|}
\hline
Method & \tabincell{c}{CO Problem} & \tabincell{c}{Model}\\ \hline\hline
ConvNet \cite{JoshiTSP} & TSP & GNN, non-autoregressive \\\hline
DTSPGNN \cite{DecisionTSP} & TSP & GNN, non-autoregressive\\ \hline
CPNGNN \cite{CPNGNN} & MDS, MM, MVC & GNN, non-autoregressive\\ \hline
GAP \cite{GAP} & Graph Partition & GNN, non-autoregressive \\ \hline
GMN \cite{GMN} & GED & GNN, non-autoregressive \\ \hline
SimGNN \cite{SimGNN} & GED & GNN, non-autoregressive \\ \hline
GRAPHSIM \cite{GRAPHSIM} & GED & GNN, non-autoregressive\\ \hline
GNNGC \cite{LemosGColor} & GColor & GNN, non-autogressive \\ \hline
SiameseGNN \cite{nowak2017note} & Graph matching, TSP & GNN, non-autogressive \\ \hline
PCAGM \cite{wang2019learning} & Graph matching & GNN, non-autogressive \\ \hline
IsoNN \cite{isoNN} & Graph Iso. & AutoEncoder, non-autogressive \\ \hline
GNNTS \cite{qifeng} & MIS, MVC, MC & GNN, non-autoregressive\\\hline
Ptr-Net \cite{pointerNN} & TSP & AutoEncoder, autoregressive \\ \hline
LSTMGMatching \cite{milan2017data} & Graph matching & AutoEncoder, autogressive \\ \hline
S2V-DQN \cite{Stru2VRL} & MVC, MaxCut, TSP & GNN, autoregressive \\ \hline
CombOptZero \cite{CombOptZero} & MVC, MaxCut, MC & GNN, autoregressive\\ \hline
RLMCS \cite{RLMCS} & MCS & GNN, autoregressive \\ \hline
CENALP \cite{JoinLinkNetAli} & Graph Alignment & SkipGram, autoregressive\\ \hline
TSPImprove \cite{ImproveTSP} & TSP & AutoEncoder, autoregressive\\ \hline
AM \cite{kool2018attention} & TSP & AutoEncoder, autoregressive \\ \hline
\end{tabular}\label{gpdlcompare}
\end{scriptsize}
\vspace{0ex}
\end{table}

\subsection{Non-autoregressive CO Methods}\label{nonauto}

Most works in this category use classification techniques to predict the class label of the nodes in the input graph. For a graph $G$, the prediction result is a $|V|\times K$ matrix ${\bf Y}$, where $K$ is the number of classes. The $u$-th row ${\bf y}_u$ of $\vec Y$ is the prediction result for the node $u$, where ${\bf y}_u[i]$ is the probability that $u$ is of the $i$-th class, for $1\leq i\leq K$. 
For example, for the minimum vertex cover (MVC) problem, the classification is binary ({\it i.e.}, $K=2$), and $\{u|{\bf y}_u[1] > {\bf y}_u[0]\}$ is the predicted solution.  For the graph partition problem, $K$ is the number of parts, and a node $u$ is classified to the part with the largest predicted probability. There are some works that predict a score for the input graphs. For example, for the graph similarity problem, the similarity score between two graphs is predicted.

\vspace{2ex}
\noindent
{\it A. Travelling Salesman Problem}
\vspace{2ex}

Joshi et al. \cite{JoshiTSP} propose a GNN-based model (ConvNet) to solve the TSP problem on Euclidean graph. The graph convolution layer of ConvNet is as follows.

\[
{\bf h}_i^{l+1} = {\bf h}_i^l + ReLU(BN({\bf W}_1^l {\bf h}_i^l + \sum_{j\in N_i}{\bf \eta}_{ij}^l\odot{\bf W}_2^l{\bf h}_j^l))
\]

\[
{\bf \eta}_{ij}^l = \frac{\sigma({\bf e}_{ij}^l)}{\sum_{j'\in N_i} \sigma({\bf e}_{ij'}^l) + \epsilon}
\]

\[
{\bf e}_{ij}^l = {\bf e}_{ij}^l + ReLU(BN({\bf W}_3^l {\bf e}_{ij}^l + {\bf W}_4^l {\bf h}_i^l + {\bf W}_5^l {\bf h}_j^l)),
\]

\noindent
where $BN$ stands for batch normalization, $\odot$ denotes element-wise product, ${\bf \eta}$ is attention weight, $\epsilon$ is a small value, ${\bf W}_1$, ${\bf W}_2$ and ${\bf W}_3$ are trainable parameters. 

The embeddings of the edges outputted by the $l$-th layer of ConvNet are fed into a multilayer perceptron (MLP) to predict $p_{ij}$ the probability of the edge $e_{ij}$ belongs to the solution of TSP. The cross entropy with the ground-truth TSP tour is used as the loss. The experiments of ConvNet show that ConvNet outperforms recent autoregressive methods but falls short of standard Operations Research solvers.

Prates et al. \cite{DecisionTSP} use GNN to solve the decision version of TSP, which is to decide if a given graph admits a Hamiltonian route with a cost no greater than a given threshold $C$. Since the weights of edges are closely related to the cost of a route, Prates et al. compute edge embedding in the graph convolution. Specifically, given a graph $G=(V,E)$, an auxiliary bipartite graph $G'=(V\cup V',E')$ is constructed, where for each edge $(u,v)$ in $G$, $G'$ has a node $n_{u,v}$ in $V'$ and edges $(n_{u,v},u)$ and $(n_{u,v},v)$ are added to $E'$. The embeddings of the nodes and edges of $G$ can be computed by a GNN on the auxiliary graph $G'$. Finally, the embeddings of the edges of $G$ are fed into an MLP to make a binary classification. If the class label of $G$ is predicted to be 1, $G$  has a Hamiltonian route with a cost no greater than $C$; otherwise, $G$ has no such route.

\vspace{2ex}
\noindent
{\it B. Graph Partition}
\vspace{2ex}

\begin{figure}
\centering
\includegraphics[width = 12cm]{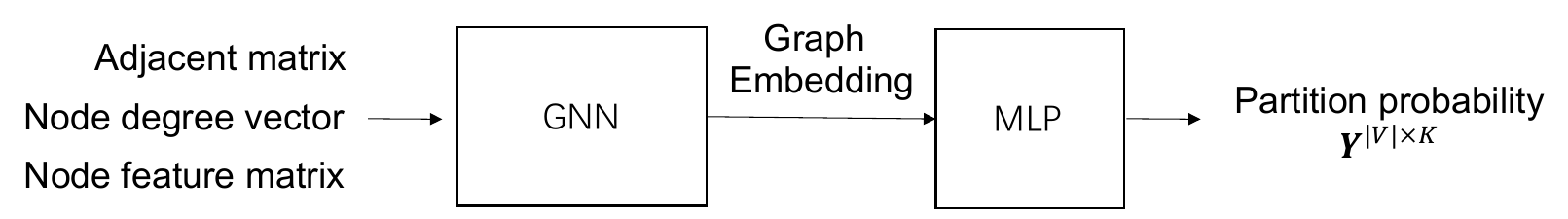}
\caption{Overview of GAP \cite{GAP}}
\label{fig:gap}
\end{figure}

Nazi et al. \cite{GAP} propose GAP as a method for computing a balanced partition of a graph. GAP is composed of  a graph embedding module, which uses a GNN model to determine the embedding of the input graph, and a graph partition module, which uses an MLP to predict the partition of nodes. The architecture of GAP is illustrated in Fig.~\ref{fig:gap}. The normalized cut size and the balancedness of the partition is used as the loss. GAP trained on a small graph can be generalized at the inference time on unseen graphs of larger size. 

Specifically, suppose $G=(V,E,{\bf X})$ is to be partitioned to $K$ disjoint parts and $V_1,V_2,...,V_K$ denote the sets of nodes of the parts, respectively.
A GNN first computes the embeddings of the nodes in $G$. Then, the MLP uses the node embeddings to predict the partition probability ${\vec Y}^{|V|\times K}$ for the nodes, where ${\vec Y}[u,i]$ is the probability that node $u$ is partitioned to $V_i$. Finally, each node  can be partitioned to the partition of the largest probability.

The loss of GAP has two components. The first component is to minimize the normalized cut size of the partition:

\[
\sum_{i=1}^K \frac{cut(V_i, \bar{V_i})}{vol(V_i)},
\]

\noindent
where $\bar{V_i}$ denotes the nodes not in $V_i$, $cut(V_i, \bar{V_i})$ denotes the number of edges crossing $V_i$ and $\bar{V_i}$, and $vol(V_i)$ denotes the total degree of the nodes in $V_i$.
The second component is to minimize the distance from the balanced
partition:

\[
\sum_{i=1}^K \sum_{u\in G} ({\bf Y}[u,i] - \frac{|V|}{K})^2,
\]

\noindent 
where $\frac{|V|}{K}$ is the part size of the balanced partition. The objective function of GAP is as follows.

\[
\min \sum_{i=1}^K \frac{cut(V_i, \bar{V_i})}{vol(V_i)} + \sum_{i=1}^K \sum_{u\in G} ({\bf Y}[u,i] - \frac{|V|}{K})^2
\]

\vspace{2ex}
\noindent
{\it C. Graph Similarity}
\vspace{2ex}

\begin{figure}
\centering
\includegraphics[width = 12cm]{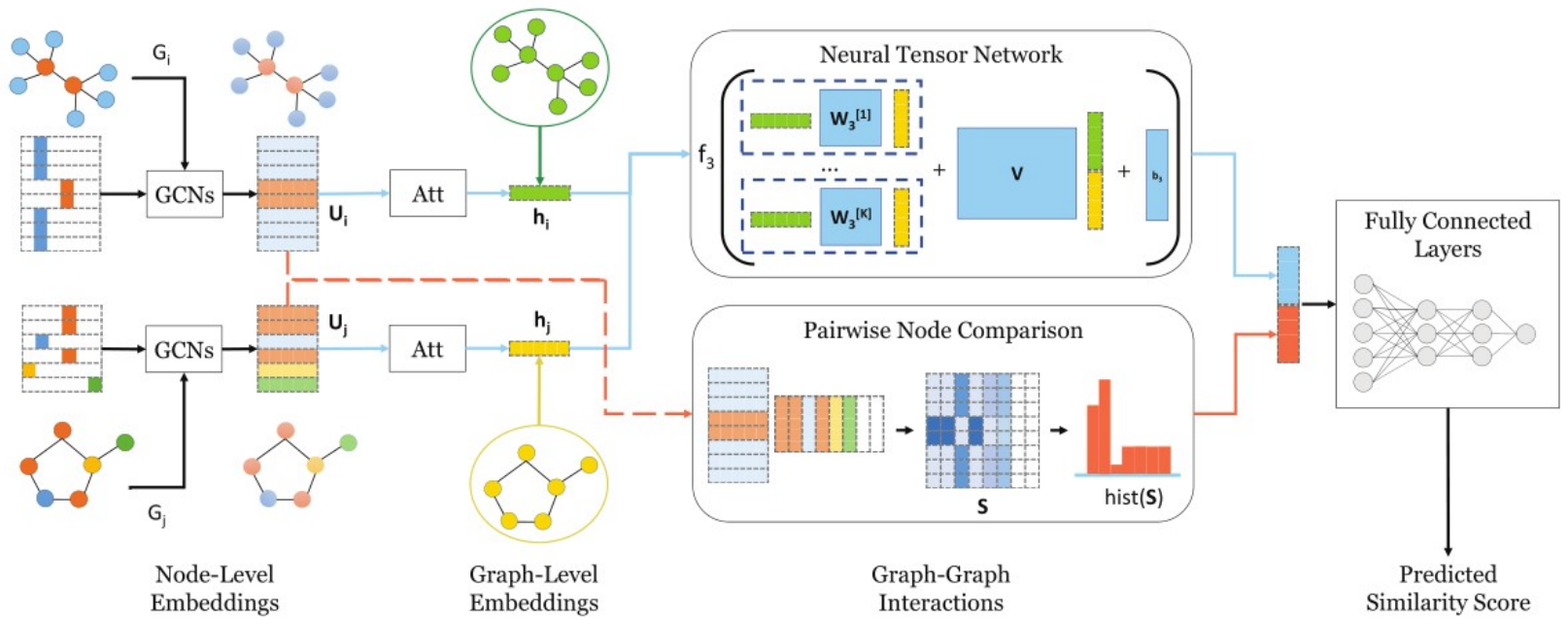}
\caption{Overview of SimGNN \cite{SimGNN}. The blue solid line illustrates the first strategy of comparing $G_1$ and $G_2$ using their global summaries ${\bf h}_{G_1}$ and ${\bf h}_{G_2}$. The orange  dashed line indicates the second strategy of the find-grained pair-wise node comparison.}
\label{fig:simgnn}
\end{figure}

Bai et al. \cite{SimGNN} propose SimGNN as a method for predicting the similarity between two graphs. SimGNN combines two strategies for predicting the similarity between two graphs $G_1$ and $G_2$. The first strategy compares $G_1$ and $G_2$ by comparing their global summaries ${\bf h}_{G_1}$ and ${\bf h}_{G_2}$. The second strategy uses the pair-wise node comparison to provide a fine-grained information as a supplement to the global summaries ${\bf h}_{G_1}$ and ${\bf h}_{G_2}$.  The architecture of SimGNN is shown in Fig.~\ref{fig:simgnn}.

As shown in Fig.~\ref{fig:simgnn}, 
SimGNN first computes the node embeddings of the two input graphs $G_1$ and $G_2$ using GCN. For the first strategy, SimGNN computes ${\bf h}_{G_1}$ and ${\bf h}_{G_2}$ from the node embeddings by means of an attention mechanism that can adaptively emphasize the important nodes with respect to a specifc similarity metric. Then, ${\bf h}_{G_1}$ and ${\bf h}_{G_2}$  are input to a neural tensor network (NTN) to compute a similarity score vector for $G_1$ and $G_2$. 

The attention mechanism to compute ${\bf h}_G$ is defined as follows. 
For a graph $G$, SimGNN introduces a context vector $\vec c$ = $tanh({\vec W} \sum_{u\in G} {\bf h}_u)$ to encode the global information of $G$. $\vec c$ is adaptive to the given similarity metric via $\vec W$. Intuitively, nodes that are close to the global context should receive more attention. Therefore, the attention weight $\alpha_u$ of a node $u$ is defined based on the inner product of $\vec c$ and ${\bf h}_u$. $\alpha_u = \sigma({\bf c}^T {\bf h}_u)$, where $\sigma$ is the sigmoid function.  The embedding of $G$, ${\bf h}_G$, is computed as ${\bf h}_G = \sum_{u\in G} \alpha_u {\bf h}_u$. 

For the second strategy, SimGNN constructs a pair-wise node similarity matrix $M$ by computing the inner product of ${\bf h}_{u}$ and ${\bf h}_{v}$ for each $u\in G_1,v\in G_2$. SimGNN uses a histogram of $M$ to summarize the pair-wise node similarity.

Finally, the similarity score vector outputted by NTN and the histogram are input to a fully connected neural network to predict the similarity between $G_1$ and $G_2$. The mean squared error between the predicted similarity with the ground truth is used as the loss of SimGNN. In the follow-up work GRAPHSIM \cite{GRAPHSIM}, a CNN-based method is used to replace the histogram of SimGNN. 

Li et al. \cite{GMN} propose the graph matching network (GMN) to solve the graph similarity problem. Instead of embedding each graph independently, GMN embeds two graphs $G_1$ and $G_2$ jointly by examining the matching between them. The matching used in GMN is soft matching, which means that a node of $G_1$ can match to all nodes of $G_2$ yet with different  strengths. The embedding of $G_1$ can change based on the other graph it is compared against. At inference time, GMN can predict if the distance between two graphs is smaller than a given threshold $\gamma$. 

Given two graphs $G_1=(V(G_1), E(G_1))$ and $G_2=(V(G_2), E(G_2))$, the $l$-th convolution layer of GMN is defined as below. 

\begin{equation}
\begin{array}{ll}
{\bf m}_{j\to i} & = MLP({\bf h}^l_i, {\bf h}^l_j), \forall (i,j)\in E(G_1) \\[0.2cm]
{\bf m}_{j'\to i'} & = MLP({\bf h}^l_{i'}, {\bf h}^l_{j'}), \forall (i',j')\in E(G_2) \\[0.2cm]
{\vec \mu}_{j'\to i} & = f_{match}({\bf h}^l_i, {\bf h}^l_{j'}), \forall i\in V(G_1), j'\in V(G_2)\\[0.2cm]
{\vec \mu}_{i\to j'} & = f_{match}({\bf h}^l_i, {\bf h}^l_{j'}), \forall i\in V(G_1), j'\in V(G_2)\\[0.2cm]
{\bf h}_i^{l+1} & = MLP({\bf h}_i^l, \sum_{j\in G_1} {\bf m}_{j\to i}, \sum_{j'\in G_2} {\bf \mu}_{j'\to i})\\[0.2cm]
{\bf h}_{j'}^{l+1} & = MLP({\bf h}_{j'}^l, \sum_{i'\in G_2} {\bf m}_{i'\to j'}, \sum_{i\in G_1} {\bf \mu}_{i\to j'}),
\end{array}
\end{equation}

\noindent
where ${\bf m}$ denotes the message aggregation of a node from its neighbors in the same graph, ${\vec \mu}$ is the cross-graph matching vector that measures the difference between a node in a graph and all the nodes in the other graph, and $f_{match}$ can be defined by the following attention based method.

\[
\begin{array}{ll}
{\vec \mu}_{j'\to i} & = \alpha_{j'\to i}({\bf h}^l_i - {\bf h}^l_{j'}), \forall i\in V(G_1), j'\in V(G_2)\\[0.2cm]
\alpha_{j'\to i} & = \frac{exp(dist({\bf h}^l_i, {\bf h}^l_{j'}))}{\sum_{v'\in G_2} exp(dist({\bf h}^l_i, {\bf h}^l_{v'}))}\\[0.2cm]
{\vec \mu}_{i\to j'} & = \alpha_{i\to j'}({\bf h}^l_i - {\bf h}^l_{j'}), \forall i\in V(G_1), j'\in V(G_2)\\[0.2cm]
\alpha_{i\to j'} & = \frac{exp(dist({\bf h}^l_i, {\bf h}^l_{j'}))}{\sum_{v\in G_1} exp(dist({\bf h}^l_{v}, {\bf h}^l_{j'}))},\\[0.2cm]
\end{array}
\]

\noindent
where $dist$ is the Euclidean distance. 

Suppose GMN stacks $L$ layers. The embedding of a graph $G$ is computed as below. 

\begin{equation}
{\bf h}_{G}  = MLP(\{{\bf h}^L_i)_{i\in G}\}),
\end{equation}

\noindent 
where ${\bf h}^L_i$ is the embedding of node $i$ outputted by the last convolution layer.

The objective function of GMN is to minimize the margin-based pairwise loss ${\mathcal L}  = \max\{0, \gamma - t \times (1-dist(G_1,G_2))\}$, 
where $\gamma>0$ is the given margin threshold, $dist(G_1,G_2)=||{\bf h}_{G_1} - {\bf h}_{G_2}||_2$ is the Euclidean distance, and $t$ is the ground truth of the similarity relationship between $G_1$ and $G_2$, {\it i.e.}, if $G_1$ and $G_2$ are similar, $t=1$; otherwise, $t=-1$.

\vspace{2ex}
\noindent
{\it D. Minimum Vertex Cover}
\vspace{2ex}

Sato et al. \cite{CPNGNN}, from a theoretical perspective, study the power of GNNs in learning approximation algorithms for the minimum vertex cover (MVC) problem. They prove that no existing GNN can compute a $(2 - \epsilon)$-approximation for MVC, where $\epsilon > 0$ is any real number and $\Delta$ is the maximum node degree.
Moreover, Sato et al. propose a more powerful consistent port numbering GNN (CPNGNN), which can return a $2$-approximation for MVC. The authors theoretically prove that there exists a set of parameters of CPNGNN that can be used to find an optimal solution for MVC. However, the authors do not propose a method for finding this set of parameters.

\begin{figure}
\centering
\includegraphics[width = 8cm]{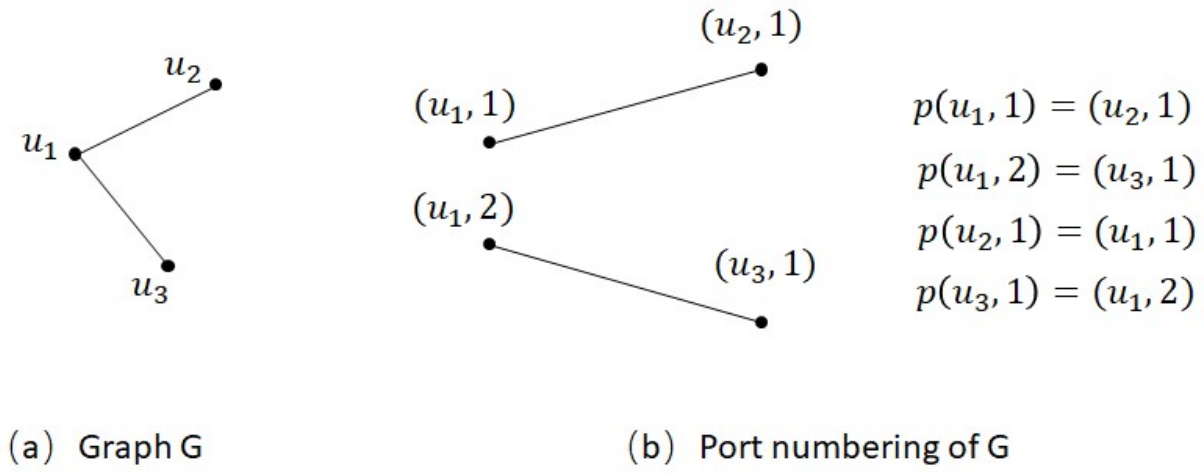}
\caption{An example of port numbering}
\label{fig:portnumb}
\end{figure}

CPNGNN is designed based on graph port numbering. Given a graph $G$, the ports of a node $u$ are pairs $(u, i)$, $1\leq i\leq |N_u|$, where $i$ is the port number. A port numbering is a function $p$ such that for any edge $(u_1,u_2)\in G$, there exists a port $(u_1, i)$ of $u_1$ and a port $(u_2, j)$ of $u_2$ satisfying $p(u_1,i)=(u_2,j)$. Intuitively, $u_1$ can send messages from the $i$th port of $u_1$ to the $j$th port of $u_2$. If $p(u_1,i)=(u_2,j)$, $u_1$ is denoted by $p_{tail}(u_2,j)$ and $i$ is denoted by $p_n(u_2,j)$. An example of port numbering is shown in Fig.~\ref{fig:portnumb}.

CPNGNN stacks $L$ convolution layers, and the $l$-th layer is defined as follows.

\begin{equation}
{\bf h}_u^l = ReLU({\bf W}^l [{\bf h}_u^{l-1} || {\bf x}_{u,1}^{l-1} || {\bf x}_{u,2}^{l-1} || ... ||{\bf x}_{u, |N_u|}^{l-1}])
\end{equation}

\[
{\bf x}_{u,i}^{l-1} = {\bf h}_{p_{tail}(u,i)}^{l-1} || p_n(u,i),
\]

\noindent
where ${\bf W}^l$ is the trainable parameter matrix and $||$ is concatenation.

Let ${\bf h}^{L}_u$ denote the embedding of $u$ outputted by the last layer of CPNGNN. An MLP takes ${\bf h}^{L}_u$ as input and outputs the prediction ${\bf y}_u$ for $u$, where ${\bf y}_u[1]$ and ${\bf y}_u[0]$ are the probabilities that $u$ is in an MVC or not, respectively. Then, the nodes $\{u|{\bf y}_u[1] > {\bf y}_u[0]\}$ are outputted as an MVC of $G$. The approximation ratio of CPNGNN is 2 for MVC.  
CPNGNN can also solve the minimum dominating set (MDS) problem and the maximum matching (MM) problem with the approximation ratio $\frac{\Delta+1}{2}$. 

\vspace{2ex}
\noindent
{\it E. Graph Coloring}
\vspace{2ex}

Lemos et al. \cite{LemosGColor} propose a graph recurrent neural network to predict if a graph is $k$-colorable. Each node $v$ has an embedding vector ${\bf h}_v$ and each color $c$ also has an embedding vector ${\bf h}_c$. Let $\vec A$ denote the adjacent matrix of the graph $G$ and $\vec M$ denote the color assignment matrix, where each row of $\vec M$ is a node of $G$ and each column of $\vec M$ is a color. ${\bf M}[v,c]=1$ means the node $v$ is assigned the color $c$. 
The embeddings of the $l+1$-th iteration ${\bf h}_v^{l+1}$ and ${\bf h}_c^{l+1}$ are computed as follows.

\[
{{\bf h}_v}^{l+1}, {{{\bf h}}_{nhid}}^{l+1} = RNN_1({{\bf h}_{nhid}}^l, {\bf A}\times {\bf h}_v^l, {\bf M}\times MLP_1({\bf h}_c^l))
\]

\[
{{\bf h}_c}^{l+1}, {{\bf h}_{chid}}^{l+1} = RNN_2({\bf h}_{chid}^l, {\bf M} \times MLP_2({\bf h}_v^l))
\]

The embeddings of nodes are fed into an MLP to predict the probability if $G$ is $k$-colorable and the loss is the binary cross entropy between the prediction and the ground-truth. Experiments of \cite{LemosGColor} show that the proposed techniques outperform the existing heuristic algorithm Tabucol and the greedy algorithm.

\vspace{2ex}
\noindent
{\it F. Graph Matching}
\vspace{2ex}

Nowak et al. \cite{nowak2017note} study the GNN-based model for the quadratic assignment problem, that can be used to address the graph matching problem. A siamese GNN is constructed to compute the embeddings of two graphs. Let $\vec Y$ be the product of the embeddings of the nodes of the two graphs. A stochastic  matrix is computed from $\vec Y$ by taking the softmax along each row (or column). The cross entropy between the stochastic matrix and the ground-truth node mapping is the loss. The proposed model can also be used to solve the TSP problem, as TSP can be formulated as a quadratic assignment problem.

Wang et al. \cite{wang2019learning} propose a GNN-based model to predict the matching of two graphs. Given two graphs $G_1$ and $G_2$, it first uses GNN to compute the embeddings of the nodes of the two graphs. Then, the embeddings are fed to a Sinkhorn layer to obtain a doubly-stochastic matrix. The cross entropy with the ground-truth node mapping is used as the loss. The idea of the Sinkhorn layer is that given a non-negative matrix, iteratively normalize each row and each column of the matrix until the sum of each row and the sum of each column equal to 1, respectively. Experiments of \cite{wang2019learning} show that the proposed model outperforms the existing learning-based graph matching methods.

\vspace{2ex}
\noindent
{\it G. Graph Isomorphism}
\vspace{2ex}

\begin{figure}
\centering
\includegraphics[width = 10cm]{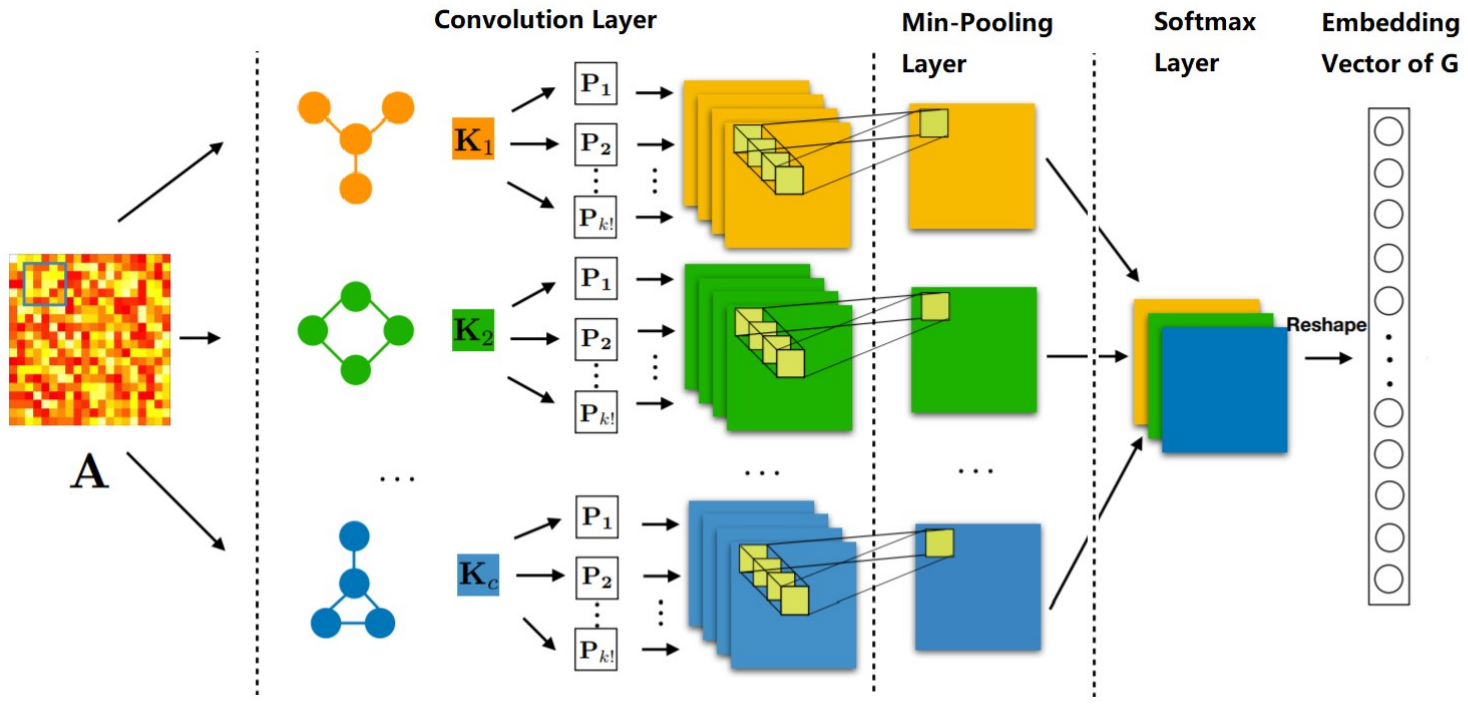}
\caption{Overview of IsoNN\cite{isoNN}}
\label{fig:isonn}
\end{figure}

Meng and Zhang \cite{isoNN} propose an isomorphic neural network (IsoNN) for learning graph embedding. The encoder  has three layers: a convolution layer, a min-pooling layer, and a softmax layer. The encoder is shown in Fig.~\ref{fig:isonn}. The decoder is an MLP to predict the binary class of $G$, and the loss is the cross entropy between the prediction and the ground truth.

Specifically, the encoder of IsoNN is designed as follows.
Given a set of motifs, the convolution layer of the encoder extracts a set of isomorphism features from $G$ for each motif. Suppose ${\bf K}_i$ is the adjacent matrix of the $i$-th motif that has $k$ nodes. The L2-norm between ${\bf K}_i$ and a $k$ by $k$ submatrix ${\bf A}_{x,y,k}$ of the adjacent matrix $\vec A$ of $G$ is an isomorphism feature extracted by ${\bf K}_i$ with respect to ${\bf A}_{x,y,k}$, where $x$ and $y$ denote the top-left corner of the submatrix in $\vec A$. IsoNN examines $k!$ permutations of ${\bf K}_i$ and extracts $k!$ isomorphism features for ${\bf A}_{x,y,k}$. The smallest one is regarded as the optimal isomorphism feature extracted by ${\bf K}_i$ for ${\bf A}_{x,y,k}$, which is computed by the min-pooling layer. Since the optimal isomorphism features for ${\bf A}_{x,y,k}$ extracted by different motifs can have different scales, the softmax layer is used to normalize them. Finally, the normalized isomorphism features extracted by all motifs for all values of $x$ and $y$ are concatenated as the embedding of $G$.

\vspace{2ex}
\noindent
{\it H. Maximum Independent Set}
\vspace{2ex}

Li et al. \cite{qifeng} propose a GNNTS model that combines GNN and heuristic search to compute the maximum independent set (MIS) of a graph. GNNTS trains a GCN $f$ using a set of training graphs, where the MISs of a graph can be used as the ground truth labels of the graph. 
For a graph $G=(V,E)$, the prediction result of $f$ is a $|V|\times 2$ matrix $\vec Y$, where ${\vec Y}[\cdot, 1]$ and ${\vec Y}[\cdot, 0]$ are the probabilities of the nodes being in or not in an MIS of $G$, respectively.

The basic idea of GNNTS is to use $f$ as the heuristic function within a greedy search procedure. Specifically, in each iteration, the nodes of $G$ are sorted by ${\bf Y}[\cdot,1]$. The greedy algorithm picks the node $u$ with the largest value in ${\bf Y}[\cdot,1]$, marks $u$ as 1, and adds $u$ to a partial solution $U$. All neighbors of $u$ are marked as 0. $u$ and its neighbors are removed from $G$, and the remaining graph is input to $f$ for the next iteration. Once all nodes in $G$ are marked, $U$ is returned as the MIS of $G$.

\begin{figure}
\centering
\includegraphics[width = 4cm]{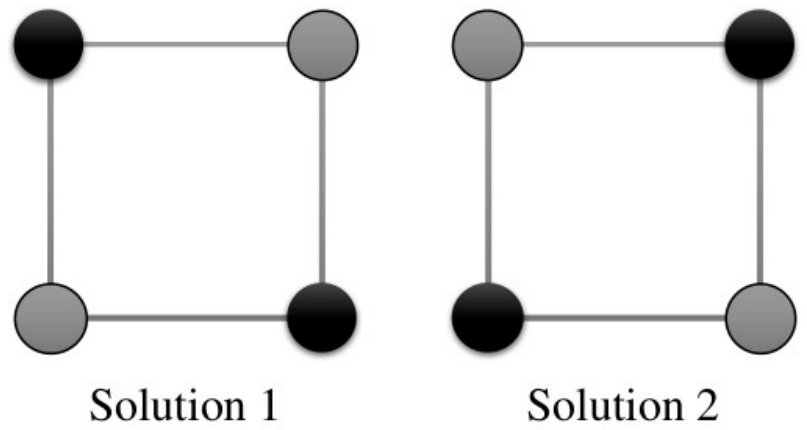}
\caption{Illustration of the two MISs of the square graph  \cite{qifeng}}
\label{fig:gnnts_case}
\end{figure}

The basic method described above has the disadvantage that it cannot support the case in which $G$ has multiple solutions. For the example shown in Fig.~\ref{fig:gnnts_case}, the square graph of four nodes has two MISs and the basic method  predicts that each node has a probability 0.5 of belonging to an MIS, which is not useful.

To address this disadvantage, the GNN $f$ is extended to output multiple prediction results, {\it i.e.}, $f(G) = \{f^1(G), f^2(G),...,f^m(G)\}$, where $f^i(G)$ is a $|V|\times 2$ matrix ${\vec Y}^i$, $1\leq i\leq m$, and $m$ is a hyperparameter. Then, the GNN $f$ is used in a tree search procedure. Specifically, GNNTS maintains a tree of partial solutions, where each leaf is a partital solution to be extended. At each step, GNNTS randomly picks a leaf $n_{leaf}$ from the search tree and uses $f$ to output $m$ prediction results ${\vec Y}^1, {\vec Y}^2, ..., {\vec Y}^m$. Then, for each ${\vec Y}^i$, GNNTS uses the basic method to compute an extension of $n_{leaf}$. The $m$ newly obtained partial solutions are inserted to the search tree as the children of $n_{leaf}$. If a leaf of the search tree cannot be extended anymore, the leaf is a maximal independent set. The largest computed maximal independent set is outputted. GNNTS can also solve the minimum vertex cover (MVC) and maximal clique (MC) problems by reducing to MIS.

\subsubsection{Discussions}

Non-autoregressive methods output a
solution in one shot. The advantage is that the inference of non-autoregressive methods is faster than autoregressive methods \cite{joshi2020learning}. However, the probability of a node/edge being a part of a solution does not depend on that of other nodes/edges. There is an opportunity that non-autoregressive methods are not able to outperform autoregressive methods for solving the CO problems having sequential characteristics, such as TSP. Therefore, there are many recent works studying autoregressive methods.

\subsection{Autoregressive CO Methods}\label{autoreg}

Autoregressive methods iteratively extend a partial solution. In each iteration, a node/edge is added to the partial solution. Most existing works use sequence model-based methods or reinforcement learning-based methods to iteratively extend the partial solution.

\vspace{1ex}
\noindent
{\it A. Sequence Model Based Methods}
\vspace{1ex}

The pointer network (Ptr-Net) proposed by Vinyals et al. \cite{pointerNN} is a seminal work in this category. It uses an RNN-based AutoEncoder to solve the travelling salesman problem (TSP) on a Euclidian graph. The encoder of Ptr-Net is an RNN taking the nodes of the graph $G$ as input and outputting an embedding of $G$, where the order of the nodes is randomly chosen. The decoder of Ptr-Net is also an RNN. In each time step, the decoder computes an attention over the input nodes, and selects the input node that has the largest attention weight as output. 

Specifically, given a graph $G$, suppose the nodes of $G$ are sequentially input as $v_1,v_2,...,v_{|V|}$ to the encoder, and the decoder sequentially outputs $v_{j_1}, v_{j_2},..., v_{j_{|V|}}$. Let ${\bf a}_1,{\bf a}_2,...,{\bf a}_{|V|}$ and ${\bf b}_1,{\bf b}_2,...,{\bf b}_{|V|}$ denote the sequences of the hidden states of the encoder and the decoder, respectively. 
For the $k$-th time step of the decoder, the decoder selects one node in $v_1,v_2,...,v_{|V|}$ as $v_{j_k}$ by an attention weight vector ${\bf \alpha}^k$ over ${\bf a}_1,{\bf a}_2,...,{\bf a}_{|V|}$. ${\bf \alpha}^k$ is defined as: 

\[
{\bf \alpha}^k[j] = {\bf c}^T [tanh ({\bf W}_1 {\bf a}_j + {\bf W}_2 {\bf b}_k)], 1\leq j\leq |V|
\]

\noindent
where ${\bf c}$, ${\bf W}_1$, and ${\bf W}_2$ are trainable parameters.
Then, the decoder outputs $v_{j_k}$ = $v_i$, where $i=argmax~{\bf \alpha}^k$.

\begin{figure}
\centering
\includegraphics[width = 10cm]{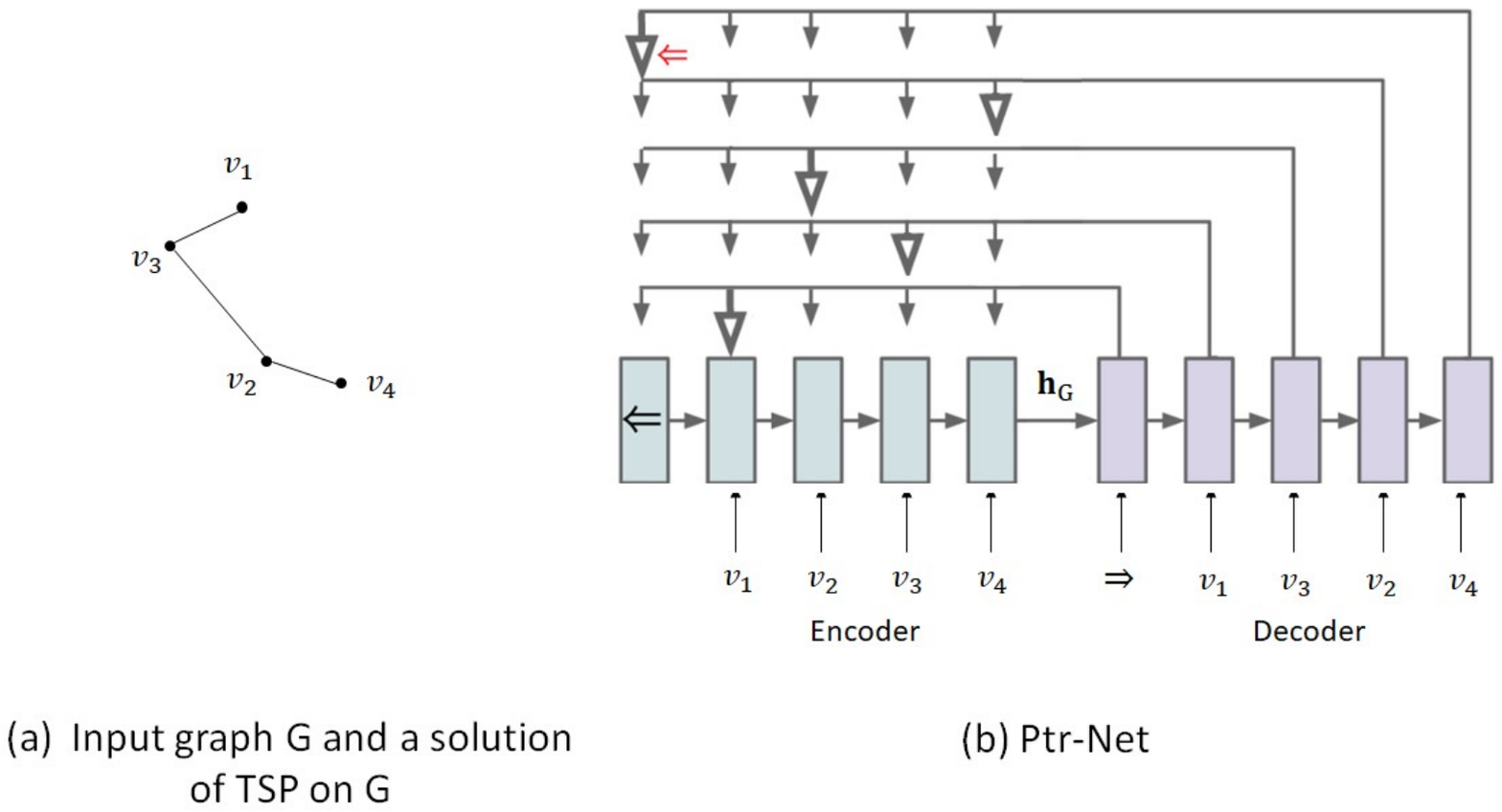}
\caption{An example of using Ptr-Net \cite{pointerNN}. (a) shows a Euclidean graph $G$ on a 2D plane, and the solution is marked by the edges. (b) shows the encoder and the decoder of Ptr-Net for finding the solution on $G$.}
\label{fig:ptrnet}
\end{figure}

For example, Fig.~\ref{fig:ptrnet}(a) shows a Euclidean graph $G$ with four nodes and a solution $v_1,v_3,v_2,v_4$. Fig.~\ref{fig:ptrnet}(b) shows the procedure of Ptr-Net for computing the solution. The hollow arrow marks the node that has the largest attention weight at each time step of the decoder.  

Milan et al. \cite{milan2017data} propose a LSTM-based method to solve the graph matching problem. 
 Given two graphs $G_1$ and $G_2$ of $n$ nodes, from the features of nodes and edges of $G_1$ and $G_2$, a $n^2$ by $n^2$ similarity matrix $\vec M$ can be computed, where ${\bf M}_{ij,lk}$ is the similarity of the edge $(v_i,v_j)\in G_1$ and $(v_j,v_k)\in G_2$, and ${\bf M}_{ii,ll}$ is the similarity of the node $v_i$ of $G_1$ and $v_l$ in $G_2$. ${\bf M}$ is input to the LSTM as the input feature. At each step, the LSTM will predict a node pair of matching. The cross entropy with the ground-truth matching is used as the loss. However, ${\bf M}$ is of $O(n^4)$ size, which is too large for matching large graphs.

Du et al. \cite{JoinLinkNetAli} observe that link prediction and graph alignment are inherently related and the joint learning of them can benefit each other. Given two graphs $G_1$ and $G_2$, crossing edges between all nodes of $G_1$ and $G_2$ are added. The network alignment model predicts the probability of accepting a crossing edge, {\it i.e.}, the end nodes of the crossing edge are aligned. The link prediction model predicts the probability of inserting an edge $(u,v)$ to $G_1$ based on if $(u',v')$ is in $G_2$, where $u$ and $v$ are aligned to $u'$ and $v'$, respectively. Both the network alignment model and the link prediction model need the embeddings of the nodes of $G_1$ and $G_2$, which are computed by the generalized SkipGram model using the random walks crossing the two graphs. Suppose the random walk is on $G_1$, it will switch to $G_2$ at the next step with probability $p$. If the random walk switches, the probability of walking from a node $v$ in $G_1$ to a node $u$ in $G_2$ is $p'(v,u)$. If the crossing edge between $v$ and $u$ is an accepted crossing edge, $p'(v,u)=1$; otherwise, $p'(v,u)=\frac{w(v,u)}{Z}$, where $w(v,u)$ is the structure similarity between $v$ and $u$ and $Z=\sum_{u'\in G_2} w(v,u')$. $w(v,u)$ is measured by the degree distributions of the neighbors of $v$ and $u$ in $G_1$ and $G_2$, respectively. In each iteration, the pair of nodes of the two graphs having the largest predicted probability by the graph alignment model is aligned and the edges of $G_1$ and $G_2$ whose probabilities predicted by the link prediction model exceed the threshold are added to $G_1$ and $G_2$, respectively. Node embeddeings are recomputed in each iteration, as the alignment between $G_1$ and $G_2$ and the edges in $G_1$ and $G_2$ are updated. Experiments of 
\cite{JoinLinkNetAli} show that link prediction and graph alignment can benefit each other and the proposed techniques are suitable for aligning graphs whose distribution of the degree of aligned nodes is close to linear or the graphs having no node attribute information.

\vspace{1ex}
\noindent
{\it B. Reinforcement Learning Based Searching}
\vspace{1ex}

When iteratively extending a partial solution, each iteration selects the node in order to optimize the final solution. Such a sequential decision process can be modeled as a Markov decision process (MDP) and solved by reinforcement learning (RL). Therefore, we first presents a brief review of RL.

\vspace{1ex}
\noindent
{\it B.1 Review of Reinforcement Learning}
\vspace{1ex}

In RL, an agent acts in an environment, collecting rewards and updating its policy to select future actions. It can be formulated as an MDP $(S, A, T, R, \gamma)$, where 
\begin{itemize}
\item $S$ is the set of states, and some states in $S$ are end states;
\item $A$ is the set of actions;
\item $T: S\times A\times S\to [0,1]$ is the transition function, $T(s,a,s')$ is the  transition probability to state $s'$ after taking action $a$ in state $s$;
\item $R: S\times A\to {\mathbb R}$ is the reward of taking action $a$ in state $s$; and
\item $\gamma$ is a discount factor.
\end{itemize}

The agent uses a policy $\pi:S\to A$ to select an action for a state. RL is to learn an optimal policy $\pi^*$ that can return the optimal action for each state in terms of the overall reward. RL relies on the state-value function and the action-value function to optimize the policy. The state-value function $V^\pi(s)$ denotes the overall reward starting from the state $s$ following the policy $\pi$. The action-value function $Q^\pi(s,a)$ denotes the overall reward starting from the state $s$ and the action $a$ following the policy $\pi$. Formally,

\[
V^\pi(s) = {\mathbb E}_\pi [\sum_{t=0}^T \gamma^t R(s_{t}, a_{t}) | s_0 = s],
\]

\[
Q^\pi(s,a) = {\mathbb E}_\pi [ \sum_{t=0}^T \gamma^k R(s_t, a_t) | s_0 = s, a_0 = a],
\]

\noindent
where ${\mathbb E}_\pi$ denotes the expected value given that the agent follows the policy $\pi$, $t$ is the time step and $T$ is the time step of reaching an ending state. The state-value function and the action-value function of the optimal policy $\pi^*$ are denoted by $V^*$ and $Q^*$, respectively.

RL can learn $\pi^*$  by iteratively optimizing the value functions, which is called as the value-based method. The value-based methods compute $Q^*$ and output the optimal policy $\pi^*(s)$ = $\max_a Q^*(s,a)$. Q-learning is a well-known value-based RL method. Suppose $Q$ is the current action-value function. At each state $s_t$, Q-learning selects the action $a_t$ by the $\epsilon$-greedy policy, which is selecting $\max_a Q(s,a)$ with a probability $1-\epsilon$ and selecting a random action with a probability $\epsilon$, and updates $Q$ as Formula~\ref{equ:qlearn}.

\begin{equation}\label{equ:qlearn}
Q(s_t,a_t) = Q(s_t,a_t) + \alpha_t[R(s_t,a_t) + \gamma\max_a Q(s_{t+1},a) - Q(s_t,a_t)],
\end{equation}

\noindent
where $\alpha_t$ is the learning rate at the time step $t$. Q-learning converges to $Q^*$ with probability 1, if each state-action pair is performed infinitely often and $\alpha_t$ satisfies $\sum_{n=1}^\infty \alpha_t = \infty$ and $\sum_{n=1}^\infty \alpha^2_t < \infty$.

Q-learning needs a table, namely Q-table, to store the action values. The size of the Q-table is $|S|\times |A|$, which can be too large to support the applications having a large number of states and actions. Therefore, many methods have been proposed to approximate the Q-table by parameterized
functions. For example, deep Q-learning network (DQN) uses  a deep neural network as the function approximation of the Q-table \cite{mnih2015human}. 

The value-based methods first optimize the value functions and then improve the policy based on the optimized value functions. There are also many methods that directly optimize the policy based on policy gradient. We refer the reader to \cite{sutton2018} for more details of RL.

\vspace{1ex}
\noindent
{\it B.2 Reinforcement Learning Based CO Methods}
\vspace{1ex}

Since iteratively extending a partial solution of a CO problem is inherently a sequential decision process, several works use reinforcement learning (RL) to extend the partial solution. The partial solution and the input graph together determine the state of RL, whereas  the node that can be added to the partial solution is the action. RL can learn an optimal policy to find the optimal node for a partial solution.

Dai et al. propose S2V-DQN \cite{Stru2VRL} that combines GNN and deep Q-learning to tackle the MVC problem. Given a graph $G$, let $U$ denote the current partial solution and $\bar{U}=V\backslash U$. The RL task for MVC can be formulated as follows.

\begin{itemize}
\item A state $s$ is determined by $G$ and $U$, $s=f_{state}(G,U)$. If $U$ is a vertex cover of $G$, the state is an end state;
\item An action $a_v$ is adding a node $v\in \bar{U}$ to $U$;
\item The transition $T(f_{state}(G,U), a_v)$ = $f_{state}(G, U\cup\{v\})$; and
\item The reward of an action $R(s,a_v) = -1$ so as to minimize the vertex cover.
\end{itemize}

The representation of state $s$ can be computed by embedding $G$ and $U$ using a GNN as follows. 

\begin{equation}\label{equ:gnn_stru2vrl}
f_{state}(G,U) = \sum_v {\bf h}^L_v
\end{equation}

\[
{\bf h}^l_u = ReLU({\vec\theta_1} x_u + {\vec \theta_2}\sum_{v\in N_u} {\bf h}_v^{l-1}+{\vec \theta_3}\sum_{v\in N_u} ReLU({\vec \theta_7} w_{u,v})),
\]

\noindent
where $L$ is the total number of layers of the GNN, $x_u=1$ if $u\in U$ and otherwise, $x_u=0$, $w_{u,v}$ is the weight of the edge $(u,v)$, and ${\vec\theta_1},{\vec\theta_2}$, and ${\vec\theta_3}$ are trainable parameters.

We can use the embedding of $v$, ${\bf h}_v$ to represent the action $a_v$. The representations of the state $s$ and the action $a_v$ are fed into an MLP to compute $Q(s,a_v)$ as below. 

\begin{equation}
Q(s,a_v) = {\vec \theta_4} ReLU(Concat({\vec \theta_5}\sum_{u\in V} {\bf h}^L_u, {\vec \theta_6}{\bf h}^L_v)),
\end{equation}

\noindent
where ${\vec\theta_4}, {\vec\theta_5}$,  and ${\vec\theta_6}$ are trainable parameters.

Deep Q-learning is used to optimize the parameters. After the MLP and the GNN are trained, they can be generalized to compute MVC for unseen graphs. S2V-DQN can also solve the MaxCut and TSP problems.

Bai et al. \cite{RLMCS} propose to compute the maximum common subgraph (MCS) of two graphs using GNN and Q-learning. Given two graphs $G_1$ and $G_2$, the partial solution is a subgraph $g_1$ of $G_1$ and a subgraph $g_2$ of $G_2$ satisfying $g_1$ and $g_2$ are isomorphic. The RL task for MCS is formulated as follows.

\begin{itemize}
\item A state $s$ is determined by $G_1$, $G_2$, $g_1$ and $g_2$, $s=f_{state}(G_1,G_2,g_1,g_2)$. If $g_1$ and $g_2$ cannot be extended, the state is an end state;
\item An action $a_{u,v}$ is to select a node $u$ from $G_1\backslash g_1$ and a node $v$ from $G_2\backslash g_2$ and add them to $g_1$ and $g_2$, respectively;
\item The transaction $T(f_{state}(G_1,G_2,g_1,g_2), a_{u,v}) = f_{state}(G_1,G_2,g_1\cup\{u\}, g_2\cup\{v\})$. The isomorphism between $g_1\cup\{u\}$ and $g_2\cup\{v\}$ needs to be assured; and
\item The reward $R(s, a_{u,v})$ = 1.
\end{itemize}

\begin{figure}
\centering
\includegraphics[width = 11cm]{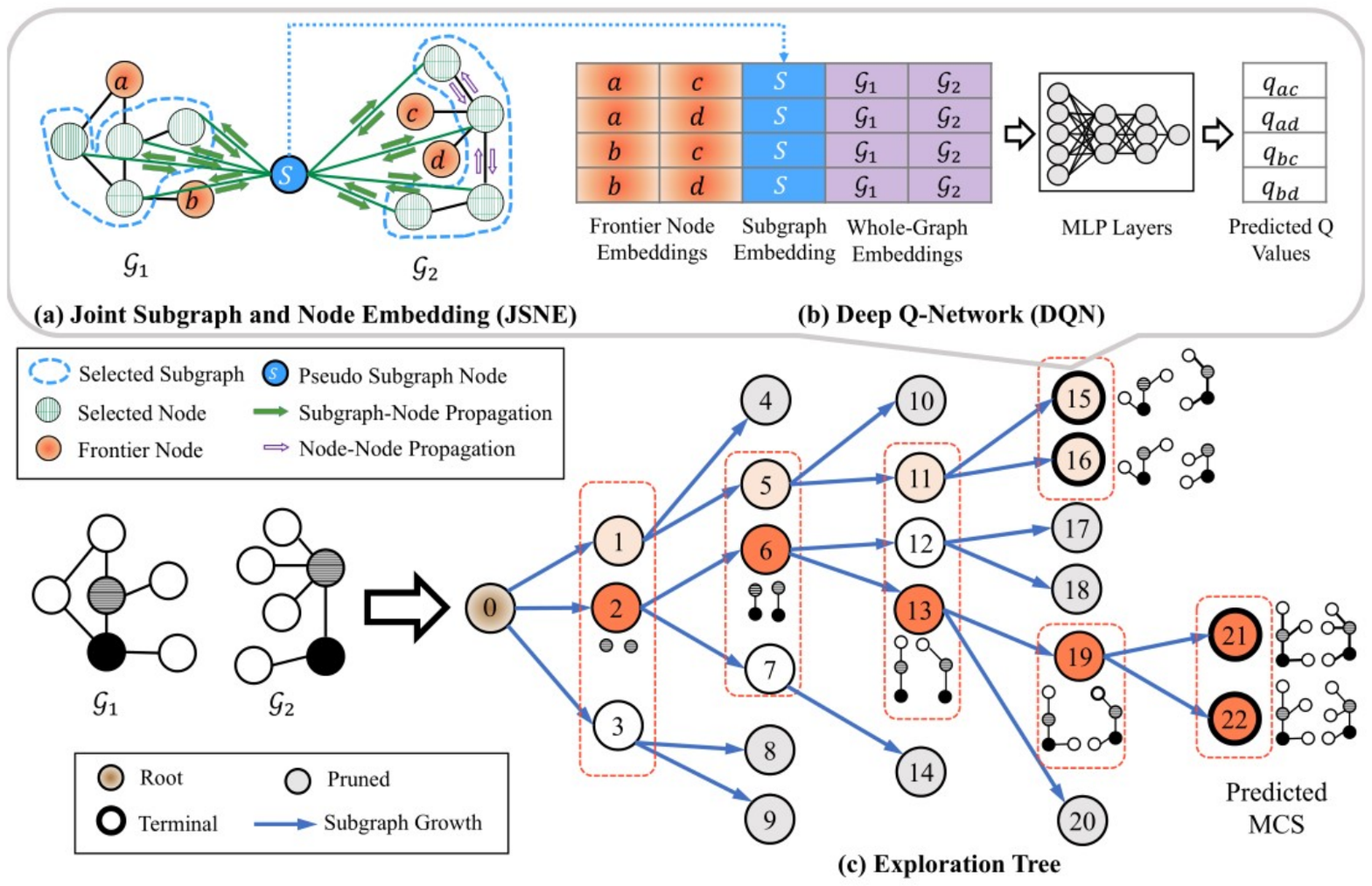}
\caption{Overview of RLMCS\cite{RLMCS}}
\label{fig:rlmcs_beam}
\end{figure}

The represention of the state $s$ can be computed  by a GNN on an auxiliary graph $G'$. $G'$ is constructed by adding a pseudo node $n_s$ connecting to the nodes in $g_1$ and the nodes in $g_2$. Then, a GNN is used to compute the node embeddings for $G'$. Note that the node embeddings change with the extension of the partial solution $g_1$ and $g_2$. ${\bf h}_{G_1}$ and ${\bf h}_{G_1}$ can be computed by the summation of the embeddings of the nodes in $G_1$ and $G_2$, respectively. The concatenation of ${\bf h}_{n_s}$, ${\bf h}_{G_1}$ and ${\bf h}_{G_1}$ is the representation of the state $s$.
The action $a_{u,v}$ is represented by the concatenation of ${\bf h}_u$ and ${\bf h}_v$. The representations of the states and the actions are fed into an MLP to predict $Q$.
Fig.~\ref{fig:rlmcs_beam}(a)-(b) show an example. 

Rather than just selecting one node with the largest Q-value as in \cite{Stru2VRL}, Bai et al. \cite{RLMCS} propose to select $k$ nodes utilizing the beam search.  At each time step, the agent of RL is allowed to transit to at most $k$ best next states. The beam search builds an exploration tree, where each node of the tree is a state and each edge of the tree is an action. Fig.~\ref{fig:rlmcs_beam}(c) shows an example of $k=3$. The partial solution is returned as a maximal independent set if it cannot be extended. The largest one among the computed maximal independent sets is outputted.

Inspired by AlphaGo Zero, which has surpassed human in the game Go, Abe et al. \cite{CombOptZero} propose CombOptZero, combining GNN and Monte Carlo tree search (MCTS)-based RL to solve the MVC problem. The formulation of the RL task is as S2V-DQN \cite{Stru2VRL}. The key difference is that CombOptZero uses the MCTS-based searching for the next action. 
For a state $s$, suppose $U$ is the partial solution, a GNN embeds $G$ and $U$ and outputs two vectors ${\bf p}$ and ${\bf v}$, where ${\bf p}[a]$ is the probability of taking the action $a$ for the state, and ${\bf v}[a]$ is the estimated overall reward from the state $s$ with action $a$. $\bf p$ and $\bf v$ are input to a MCTS, which can produce a better action prediction ${\bf p}'$ than $\bf p$. $argmax_a~{\bf p}'[a]$ is outputted as the optimal action selected for $s$. CombOptZero can also solve the MaxCut problem.

Kool et al. \cite{kool2018attention} use AutoEncoder to sequentially output a TSP tour of a graph $G$. 
The encoder stacks $L$ self-attention layers. The details of the encoder are presented in Sec.~\ref{end2end_autoencoder}.

The decoder of \cite{kool2018attention} sequentially predicts the next node to be added to the partial solution $seq$, {\it i.e.}, a partial TSP tour. At the $t$-th step of decoding, $seq$ has $t-1$ nodes. A special context vector ${\bf h}_c$ is introduced to represent the decoding context. At the $t$-th step of decoding,  ${\bf h}_c = {\bf h}_G || {\bf h}_{seq_{t-1}} || {\bf h}_{seq_0}$, where $||$ denotes concatenation, $seq_0$ denotes the 0-th node in $seq$ and $seq_{t-1}$ denotes the $t-1$-th node in $seq$. The embedding of a node $v_i$ is computed as ${\bf h}_{v_i}=\sum_{j\in N_i} \alpha_j {\bf W}_1 {\bf h}_j$, where the attention weight $\alpha_j=\frac{e^{u_j}}{\sum_{v_{j'}\in N_i} e^{u_{j'}}}$ and $u_j=({\bf W}_2{\bf h}_c)^T ({\bf W}_3{\bf h}_{v_j})$, if $v_j\not\in seq$; otherwise, $u_j=-\infty$. The probability of choosing $v_i$ to add to $seq$ at the $t$-th step is $p_{v_i}=\frac{e^{u_j}}{\sum_{v_{j'}\in G} e^{u_{j'}}}$. ${\bf W}_1, {\bf W}_2$ and ${\bf W}_3$ are trainable parameters. The REINFORCE algorithm is used to train the model. 

The experiments presented in \cite{kool2018attention} show that the proposed method can support several related problems of TSP, including Vehicle Routing Problem (VRP), Orienteering Problem (OP), Prize Collecting TSP (PCTSP) and Stochastic PCTSP (SPCTSP) with the same set of hyperparameters. However, the proposed method does not outperform the specialized algorithm for TSP ({\it e.g.}, Concorde).

There are works not iteratively extending a partial solution to a solution of a CO problem but iteratively improving a suboptimal solution to a better solution. For example, 
Wu et al. \cite{ImproveTSP} propose to improve the solution of TSP ({\it i.e.}, a TSP tour) on $G$ using RL. The MDP is defined as follows. A TSP tour of $G$ is a state $s=(v_1, v_2, ..., v_n)$, $n$ is the number of nodes in $G$ and $v_i\neq v_j$ for $i\neq j$. A 2-opt operator is an action. Given two nodes $v_i,v_j$ in $s$, the 2-opt operator selects a pair of nodes $v_i$ and $v_j$ and reverses the order of nodes between $v_i$ and $v_j$ in $s$. The transition of an action is deterministic. The reward of an action is the reduction of the TSP tour with respect to the current best TSP tour so far. The architecture of Transformer is adopted to compute node embeddings. The compatibility of a pair of nodes $v_i$ and $v_j$ is computed as $({\bf W}_1 {\bf h}_i)^T ({\bf W}_2{\bf h}_j)$, where ${\bf W}_1$ and ${\bf W}_2$ are trainable parameters. The compatibilities of all pairs of nodes are stored in a matrix $\vec Y$. $\vec Y$ is fed into a masked softmax layer as follows.

\[
{\bf Y}'_{i,j}=\left\{\begin{array}{ll}
C\cdot tanh(Y_{i,j}), & \text{if $i\neq j$}\\
-\infty, & \text{if $i=j$}
\end{array}\right.
\]

\[
P = softmax({\bf Y}'),
\]

\noindent
where $P_{i,j}$ is the probability of selecting the pair of nodes $v_i$ and $v_j$ in the 2-opt operator. REINFORCE is used to train the model. The experiments reported in \cite{ImproveTSP} show that the proposed techniques outperform the heuristic algorithms for improving TSP tours.

\subsection{Discussions}

Non-autoregressive methods predict the probabilities that each node/edge being a part of a solution in one shot. The cross entropy between the predicted probabilities and the ground-truth solution of the CO problem is used as the loss function. Autoregressive methods predict the node/edge to add to the partial solution step by step. The inference of non-autoregressive methods can be faster than autoregressive methods, as when performing inference non-autoregressive methods predict a solution in one shot. Fast inference is desired for some real-time decision-making tasks, {\it e.g.}, the vehicle routing problem.  However, non-autoregressive methods inherently ignore some sequential characteristics of some CO problems, {\it e.g.}, the TSP problem. Autoregressive methods can explicitly model this sequential inductive bias by attention mechanism or recurrent neural networks. Experimental comparison in \cite{joshi2020learning} shows that the autoregressive methods can outperform the non-autoregressive methods in terms of the quality of the tour found for the TSP problem but takes much longer time. However, for the problem without sequential characteristic, non-autoregressive methods can produce better solution, {\it e.g.}, molecule generation task \cite{pmlr-v80-jin18a}.

The non-autoregressive methods need the ground-truth solution for supervised training. It is a drawback of the non-autoregressive methods as it is hard to compute the ground-truth solution for the CO problems on large graphs, considering the NP-hardness of the CO problems. The autoregressive methods with reinforcement learning-based searching do not need the ground-truth, which has the potential to support larger graphs. Moreover, the supervised learning of non-autoregressive models that having a large number of parameters can make the models remember the training instances and the generalization is limited on unseen instances. Although reinforcement learning can overcome this problem, the sample efficiency needs to be improved. 

Regarding the comparison with traditional heuristic algorithms for the CO problems, current learning-based CO methods can have competitive performance with the traditional problem-specific heuristic algorithms on small graphs, but current learning-based CO methods do not scale well to large graphs. As large graphs have been emerging in many applications, there have been a trend of studying learning-based methods on large graphs.

The techniques and ideas of traditional heuristics for the CO problems can benefit the learning-based CO methods. For example, Dai et al. \cite{Stru2VRL} present that incorporating the idea of adding the farthest nodes first and the 2-opt operation of traditional heuristics can improve the performance of the learning-based method for TSP. Exploring the chance of integrating the ideas and operations of traditional heuristics into the learning-based methods is attracting increasing research attention \cite{deudon2018learning, Stru2VRL, kool2018attention}.

\section{Future Work}\label{future}

 Although there are recent significant advances of using graph learning models in solving several different CO problems, graph learning-based methods for CO problems are still at the early stage and there are many open problems for further studies. Some possible directions are listed as follows.

{\bf Encoding global information of graphs.} In many graph-based CO problems, the global information of the graph is needed for solving the CO problem ({\it e.g.}, graph edit distance, TSP). However, existing graph learning models, especially graph convolution, is aggregating local information from neighbors. Although more global information can be obtained by adding more graph convolution layers, there may be a non-trivial  over-smooth problem. Therefore, how to effectively encode more global information of graphs is an important direction.

{\bf Designing task-dependent model.} A GNN architecture is used to support diverse types of CO problems. However, each problem has its own characteristics. How to encode inductive bias into GNN architectures in order to better capture the characteristics of the CO problems is an important direction. 

The loss function that is generally used in classification or regression ({\it e.g.}, cross entropy) is widely used in the learning-based methods for solving CO problems. However, the general loss function may not have a strong relationship with the objective of the CO problems. For example, switching two nodes in a TSP tour will produce a TSP tour of very different score with respect to the objective of TSP. However, the two TSP tours can have the same loss in terms of cross entropy \cite{milan2017data}. Therefore, designing the problem-specific loss function needs to be studied.

{\bf Generalization.} Most existing learning-based methods for a CO problem cannot outperform traditional heuristic algorithms specifically designed for the CO problem on a larger graph or the graphs unseen in  training, although the learning-based methods can be on par with or better than the traditional heuristic algorithm on small graphs. Therefore, an important direction is to rethink the learning pipeline for CO
problem in order to generalize to larger graphs and unseen graphs \cite{joshi2020learning}.

{\bf Integration of traditional heuristics.} Integrating traditional heuristics can improve the performance of learning-based CO methods. For example, Dai et al. \cite{Stru2VRL} present that incorporating the idea of adding the farthest nodes first and the 2-opt operation of traditional heuristics can improve the performance of the learning-based method for TSP. Therefore, identifying the operations of traditional heuristics of a CO problem that can benefit the learning-based methods for the CO problem and integrating the operations appropriately into the learning procedure need to be  studied.

{\bf Supporting many graphs.} Most existing graph learning based CO methods focus on a graph or two graphs. 
Another possible future direction is to study the problems that involve a large number of graphs, for example, by optimizing the query evaluation on a large graph database such as graph similarity search, graph pattern matching query and subgraph isomorphism search.

\section{Conclusion} \label{conc}

In this survey, we provided a thorough overview of the recent graph learning methods for solving CO problems. Existing works fall into two main categories. First, non-autoregressive methods predict the solution of a CO problem in one shot. Second, autoregressive methods iterative extend a partial solution step by step. Heuristic search and reinforcement learning are widely used in the autoregressive methods to extend the partial solution. In these graph learning based CO methods, a graph is represented in numerical vectors. Then, we also survey the recent graph representation learning methods, including the generalized SkipGram-based methods, the AutoEncoder-based methods and the GNN-based methods. Several possible directions for future research are discussed as well.

\section{List of abbreviations}\label{sec-abbr}

ML, machine learning; GNN, graph neural network; DL, deep learning; RL, reinforcement learning; CNN, convolutional neural network; DNN, deep neural
network; RNN, recurrent neural network; MLP, multi-layer perceptron; MDP, Markov decision process; MCTS, Monte Carlo tree search; CO, combinatorial optimization; MVC, minimum vertex cover; MIS, maximum independent set; TSP, travelling salesman problem; GC, graph coloring; MDS, minimum dominating set; MM, maximum matching; MaxCut, maximum cut; MC, maximum clique; SI, subgraph isomorphism; GSim, graph similarity; MF, matrix factorization; B\&B, branch and bound, MILP, mixed-integer linear programming; BFS, breadth-first search; DFS, depth-first search.

\bibliographystyle{spmpsci}      
\bibliography{template}   

\begin{thebibliography}{10}
\providecommand{\url}[1]{{#1}}
\providecommand{\urlprefix}{URL }
\expandafter\ifx\csname urlstyle\endcsname\relax
  \providecommand{\doi}[1]{DOI~\discretionary{}{}{}#1}\else
  \providecommand{\doi}{DOI~\discretionary{}{}{}\begingroup
  \urlstyle{rm}\Url}\fi

\bibitem{CombOptZero}
Abe, K., Xu, Z., Sato, I., Sugiyama, M.: Solving np-hard problems on graphs
  with extended alphago zero.
\newblock arXiv \textbf{1905.11623} (2019)

\bibitem{SimGNN}
Bai, Y., Ding, H., Bian, S., Chen, T., Sun, Y., Wang, W.: {S}im{GNN}: A neural
  network approach to fast graph similarity computation.
\newblock In: Proceedings of the ACM International Conference on Web Search and
  Data Mining (WSDM'19), pp. 384--392 (2019)

\bibitem{GRAPHSIM}
Bai, Y., Ding, H., Gu, K., Sun, Y., Wang, W.: Learning-based efficient graph
  similarity computation via multi-scale convolutional set matching.
\newblock In: Proceedings of the AAAI Conference on Artificial Intelligence
  (AAAI'20), pp. 3219--3226 (2020)

\bibitem{RLMCS}
Bai, Y., Xu, D., Wang, A., Gu, K., Wu, X., Marinovic, A., Ro, C., Sun, Y.,
  Wang, W.: Fast detection of maximum common subgraph via deep {Q}-learning.
\newblock arXiv \textbf{2002.03129} (2020)

\bibitem{MILP}
Bengio, Y., Lodi, A., Prouvost, A.: Machine learning for combinatorial
  optimization: A methodological tour d'horizon.
\newblock arXiv \textbf{1811.06128} (2018)

\bibitem{bonner2019exploring}
Bonner, S., Kureshi, I., Brennan, J., Theodoropoulos, G., McGough, A.S., Obara,
  B.: Exploring the semantic content of unsupervised graph embeddings: An
  empirical study.
\newblock Data Science and Engineering \textbf{4}(3), 269--289 (2019)

\bibitem{CaiSurvey}
Cai, H., Zheng, V.W., Chang, K.: A comprehensive survey of graph embedding:
  Problems, techniques, and applications.
\newblock IEEE Transactions on Knowledge and Data Engineering \textbf{30}(09),
  1616--1637 (2018)

\bibitem{GraRep}
Cao, S., Lu, W., Xu, Q.: {G}ra{R}ep: Learning graph representations with global
  structural information.
\newblock In: Proceedings of ACM International on Conference on Information and
  Knowledge Management (CIKM'15), pp. 891--900 (2015)

\bibitem{chamiSurvey}
Chami, I., Abu-El-Haija, S., Perozzi, B., Ré, C., Murphy, K.: Machine learning
  on graphs: A model and comprehensive taxonomy.
\newblock arXiv \textbf{2005.03675} (2020)

\bibitem{MCsparse}
Chang, L.: Efficient maximum clique computation over large sparse graphs.
\newblock In: Proceedings of ACM SIGKDD International Conference on Knowledge
  Discovery and Data Mining (KDD'19), pp. 529--538 (2019)

\bibitem{chenSurvey}
Chen, F., Wang, Y.C., Wang, B., Kuo, C.C.J.: Graph representation learning: A
  survey.
\newblock APSIPA Transactions on Signal and Information Processing \textbf{9},
  e15 (2020)

\bibitem{GraphCSC}
{Chen}, H., {Yin}, H., {Chen}, T., {Nguyen}, Q.V.H., {Peng}, W., {Li}, X.:
  Exploiting centrality information with graph convolutions for network
  representation learning.
\newblock In: IEEE International Conference on Data Engineering (ICDE'19), pp.
  590--601 (2019)

\bibitem{CuiPeng}
{Cui}, P., {Wang}, X., {Pei}, J., {Zhu}, W.: A survey on network embedding.
\newblock IEEE Transactions on Knowledge and Data Engineering \textbf{31}(5),
  833--852 (2019)

\bibitem{Stru2VRL}
Dai, H., Khalil, E.B., Zhang, Y., Dilkina, B., Song, L.: Learning combinatorial
  optimization algorithms over graphs.
\newblock In: Proceedings of International Conference on Neural Information
  Processing Systems (NIPS’17), pp. 6351--6361 (2017)

\bibitem{motif2vec}
{Dareddy}, M.R., {Das}, M., {Yang}, H.: Motif2{V}ec: Motif aware node
  representation learning for heterogeneous networks.
\newblock In: 2019 IEEE International Conference on Big Data (Big Data'19), pp.
  1052--1059 (2019)

\bibitem{dave2019neural}
Dave, V.S., Zhang, B., Chen, P.Y., Al~Hasan, M.: Neural-brane: Neural bayesian
  personalized ranking for attributed network embedding.
\newblock Data Science and Engineering \textbf{4}(2), 119--131 (2019)

\bibitem{deudon2018learning}
Deudon, M., Cournut, P., Lacoste, A., Adulyasak, Y., Rousseau, L.M.: Learning
  heuristics for the {TSP} by policy gradient.
\newblock In: International conference on the integration of constraint
  programming, artificial intelligence, and operations research, pp. 170--181
  (2018)

\bibitem{devlin-etal-2019-bert}
Devlin, J., Chang, M.W., Lee, K., Toutanova, K.: {BERT}: Pre-training of deep
  bidirectional transformers for language understanding.
\newblock In: Proceedings of Conference of the North {A}merican Chapter of the
  Association for Computational Linguistics: Human Language Technologies,
  Volume 1, pp. 4171--4186 (2019)

\bibitem{JoinLinkNetAli}
Du, X., Yan, J., Zha, H.: Joint link prediction and network alignment via
  cross-graph embedding.
\newblock In: Proceedings of International Joint Conference on Artificial
  Intelligence, {IJCAI'19}, pp. 2251--2257 (2019)

\bibitem{Fingerprints}
Duvenaud, D.K., Maclaurin, D., Iparraguirre, J., Bombarell, R., Hirzel, T.,
  Aspuru-Guzik, A., Adams, R.P.: Convolutional networks on graphs for learning
  molecular fingerprints.
\newblock In: Advances in Neural Information Processing Systems 28, pp.
  2224--2232 (2015)

\bibitem{GNNRec}
Fan, W., Ma, Y., Li, Q., He, Y., Zhao, E., Tang, J., Yin, D.: Graph neural
  networks for social recommendation.
\newblock In: The World Wide Web Conference, pp. 417--426 (2019)

\bibitem{subisoAuth}
{Fan}, Z., {Peng}, Y., {Choi}, B., {Xu}, J., {Bhowmick}, S.S.: Towards
  efficient authenticated subgraph query service in outsourced graph databases.
\newblock IEEE Transactions on Services Computing \textbf{7}(4), 696--713
  (2014)

\bibitem{maxcutNP}
Goemans, M.X., Williamson, D.P.: Improved approximation algorithms for maximum
  cut and satisfiability problems using semidefinite programming.
\newblock Journal of the ACM \textbf{42}(6), 1115--1145 (1995)

\bibitem{node2vec}
Grover, A., Leskovec, J.: Node2{V}ec: Scalable feature learning for networks.
\newblock In: Proceedings of ACM SIGKDD International Conference on Knowledge
  Discovery and Data Mining (KDD’16), pp. 855--864 (2016)

\bibitem{GraphSAGE}
Hamilton, W.L., Ying, R., Leskovec, J.: Inductive representation learning on
  large graphs.
\newblock In: Proceedings of International Conference on Neural Information
  Processing Systems (NIPS’17), pp. 1025--1035 (2017)

\bibitem{HamiltonSurvey}
Hamilton, W.L., Ying, R., Leskovec, J.: Representation learning on graphs:
  Methods and applications.
\newblock IEEE Data Engineering Bulletin  (2017)

\bibitem{MCInappro}
H{\aa}stad, J.: Clique is hard to approximate within $n^{1-\varepsilon}$.
\newblock Acta Mathematica \textbf{182}(1), 105--142 (1999)

\bibitem{HopfieldNNTSP}
Hopfield, J., Tank, D.: Neural computation of decisions in optimisation
  problems.
\newblock Biological Cybernetics \textbf{52}, 141--152 (1985)

\bibitem{GCBetter}
Huang, J., Patwary, M.M.A., Diamos, G.F.: Coloring big graphs with alphagozero.
\newblock CoRR \textbf{abs/1902.10162} (2019)

\bibitem{ASGCN}
Huang, W., Zhang, T., Rong, Y., Huang, J.: Adaptive sampling towards fast graph
  representation learning.
\newblock In: Proceedings of International Conference on Neural Information
  Processing Systems (NIPS’18), pp. 4563--4572 (2018)

\bibitem{hlx2019community}
Huang, X., Lakshmanan, L.V., Xu, J.: Community Search over Big Graphs.
\newblock Morgan \& Claypool Publishers (2019)

\bibitem{pmlr-v80-jin18a}
Jin, W., Barzilay, R., Jaakkola, T.: Junction tree variational autoencoder for
  molecular graph generation.
\newblock In: Proceedings of International Conference on Machine Learning,
  vol.~80, pp. 2323--2332 (2018)

\bibitem{joshi2020learning}
Joshi, C.K., Cappart, Q., Rousseau, L.M., Laurent, T., Bresson, X.: Learning
  {TSP} requires rethinking generalization.
\newblock arXiv \textbf{2006.07054} (2020)

\bibitem{JoshiTSP}
Joshi, C.K., Laurent, T., Bresson, X.: An efficient graph convolutional network
  technique for the travelling salesman problem.
\newblock CoRR \textbf{abs/1906.01227} (2019)

\bibitem{GCInappro}
{Khot}, S.: Improved inapproximability results for maxclique, chromatic number
  and approximate graph coloring.
\newblock In: Proceedings IEEE Symposium on Foundations of Computer Science,
  pp. 600--609 (2001)

\bibitem{GCN}
Kipf, T.N., Welling, M.: Semi-supervised classification with graph
  convolutional networks.
\newblock In: International Conference on Learning Representations (ICLR'17)
  (2017)

\bibitem{kool2018attention}
Kool, W., van Hoof, H., Welling, M.: Attention, learn to solve routing
  problems!
\newblock In: International Conference on Learning Representations (2019)

\bibitem{LambNeuralSymbolic}
Lamb, L.C., Garcez, A.d., Gori, M., Prates, M.O., Avelar, P.H., Vardi, M.Y.:
  Graph neural networks meet neural-symbolic computing: A survey and
  perspective.
\newblock In: Proceedings of International Joint Conference on Artificial
  Intelligence (IJCAI'20), pp. 4877--4884 (2020)

\bibitem{w2c}
Le, Q., Mikolov, T.: Distributed representations of sentences and documents.
\newblock In: Proceedings of International Conference on International
  Conference on Machine Learning - Volume 32, ICML’14, p.
  II–1188–II–1196 (2014)

\bibitem{LemosGColor}
Lemos, H., Prates, M.O.R., Avelar, P.H.C., Lamb, L.C.: Graph colouring meets
  deep learning: Effective graph neural network models for combinatorial
  problems.
\newblock CoRR \textbf{abs/1903.04598} (2019)

\bibitem{GMN}
Li, Y., Gu, C., Dullien, T., Vinyals, O., Kohli, P.: Graph matching networks
  for learning the similarity of graph structured objects.
\newblock In: Proceedings of International Conference on Machine Learning
  (ICML'19), pp. 3835--3845 (2019)

\bibitem{qifeng}
Li, Z., Chen, Q., Koltun, V.: Combinatorial optimization with graph
  convolutional networks and guided tree search.
\newblock In: Proceedings of International Conference on Neural Information
  Processing Systems (NIPS’18), pp. 537--546 (2018)

\bibitem{RLCO}
Mazyavkina, N., Sviridov, S., Ivanov, S., Burnaev, E.: Reinforcement learning
  for combinatorial optimization: A survey.
\newblock arXiv \textbf{2003.03600} (2020)

\bibitem{isoNN}
Meng, L., Zhang, J.: {IsoNN}: Isomorphic neural network for graph
  representation learning and classification.
\newblock CoRR \textbf{abs/1907.09495} (2019)

\bibitem{w2v}
Mikolov, T., Sutskever, I., Chen, K., Corrado, G., Dean, J.: Distributed
  representations of words and phrases and their compositionality.
\newblock In: Proceedings of the 26th International Conference on Neural
  Information Processing Systems (NIPS'13), pp. 3111--3119 (2013)

\bibitem{milan2017data}
Milan, A., Rezatofighi, S.H., Garg, R., Dick, A., Reid, I.: Data-driven
  approximations to np-hard problems.
\newblock In: Thirty-First AAAI Conference on Artificial Intelligence (2017)

\bibitem{mnih2015human}
Mnih, V., Kavukcuoglu, K., Silver, D., Rusu, A.A., Veness, J., Bellemare, M.G.,
  Graves, A., Riedmiller, M., Fidjeland, A.K., Ostrovski, G., et~al.:
  Human-level control through deep reinforcement learning.
\newblock Nature \textbf{518}(7540), 529--533 (2015)

\bibitem{subgraph2vec}
Narayanan, A., Chandramohan, M., Chen, L., Liu, Y., Saminathan, S.:
  Subgraph2{V}ec: Learning distributed representations of rooted sub-graphs
  from large graphs

\bibitem{GAP}
Nazi, A., Hang, W., Goldie, A., Ravi, S., Mirhoseini, A.: {GAP} : Generalizable
  approximate graph partitioning framework (2019).
\newblock \urlprefix\url{ICLR workshop}

\bibitem{MGCN}
Nguyen, H., Murata, T.: Motif-aware graph embeddings.
\newblock In: Proceedings of International Joint Conference on Artificial
  Intelligence (IJCAI'17), pp. 1--1 (2017)

\bibitem{nowak2017note}
Nowak, A., Villar, S., Bandeira, A.S., Bruna, J.: A note on learning algorithms
  for quadratic assignment with graph neural networks.
\newblock In: Proceeding of International Conference on Machine Learning
  (ICML'17), pp. 22--22 (2017)

\bibitem{HOPE}
Ou, M., Cui, P., Pei, J., Zhang, Z., Zhu, W.: Asymmetric transitivity
  preserving graph embedding.
\newblock In: Proceedings of ACM SIGKDD International Conference on Knowledge
  Discovery and Data Mining (KDD'16), pp. 1105--1114 (2016)

\bibitem{tspInapproximable}
Papadimitriou, C.H., Vempala†, S.: On the approximability of the traveling
  salesman problem.
\newblock Combinatorica \textbf{26}(1), 101--120 (2006)

\bibitem{VColor}
{Peng}, Y., {Choi}, B., {He}, B., {Zhou}, S., {Xu}, R., {Yu}, X.: {VC}olor: A
  practical vertex-cut based approach for coloring large graphs.
\newblock In: IEEE International Conference on Data Engineering (ICDE'16), pp.
  97--108 (2016)

\bibitem{authSubSim}
{Peng}, Y., {Fan}, Z., {Choi}, B., {Xu}, J., {Bhowmick}, S.S.: Authenticated
  subgraph similarity searchin outsourced graph databases.
\newblock IEEE Transactions on Knowledge and Data Engineering \textbf{27}(7),
  1838--1860 (2015)

\bibitem{deepwalk}
Perozzi, B., Al-Rfou, R., Skiena, S.: {D}eep{W}alk: Online learning of social
  representations.
\newblock In: Proceedings of ACM SIGKDD International Conference on Knowledge
  Discovery and Data Mining (KDD'14), pp. 701--710 (2014)

\bibitem{DecisionTSP}
Prates, M., Avelar, P., Lemos, H., Lamb, L., Vardi, M.: Learning to solve
  {NP}-{C}omplete problems - a graph neural network for decision {TSP}.
\newblock In: Proceedings of {AAAI} Conference on Artificial Intelligence
  (AAAI'19), pp. 4731--4738 (2019)

\bibitem{Struc2vec}
Ribeiro, L.F., Saverese, P.H., Figueiredo, D.R.: Struc2{V}ec: Learning node
  representations from structural identity.
\newblock In: Proceedings of ACM SIGKDD International Conference on Knowledge
  Discovery and Data Mining (KDD'17), pp. 385--394 (2017)

\bibitem{CPNGNN}
Sato, R., Yamada, M., Kashima, H.: Approximation ratios of graph neural
  networks for combinatorial problems.
\newblock In: Proceedings of the Neural Information Processing Systems
  (NIPS'19) (2019)

\bibitem{Smith99}
Smith-Miles, K.: Neural networks for combinatorial optimization: A review of
  more than a decade of research.
\newblock INFORMS Journal on Computing \textbf{11}, 15--34 (1999)

\bibitem{sutton2018}
Sutton, R.S., Barto, A.G.: Reinforcement learning: An introduction.
\newblock MIT press (2018)

\bibitem{PTE}
Tang, J., Qu, M., Mei, Q.: {PTE}: Predictive text embedding through large-scale
  heterogeneous text networks.
\newblock In: Proceedings of ACM SIGKDD International Conference on Knowledge
  Discovery and Data Mining (KDD’15), pp. 1165--1174 (2015)

\bibitem{LINE}
Tang, J., Qu, M., Wang, M., Zhang, M., Yan, J., Mei, Q.: {LINE}: Large-scale
  information network embedding.
\newblock In: Proceedings of International Conference on World Wide Web
  (WWW'15), pp. 1067--1077 (2015)

\bibitem{VERSE}
Tsitsulin, A., Mottin, D., Karras, P., M\"{u}ller, E.: {VERSE}: Versatile graph
  embeddings from similarity measures.
\newblock In: Proceedings of World Wide Web Conference (WWW'18), pp. 539--548
  (2018)

\bibitem{DeepRecurNN}
Tu, K., Cui, P., Wang, X., Yu, P.S., Zhu, W.: Deep recursive network embedding
  with regular equivalence.
\newblock In: Proceedings of ACM SIGKDD International Conference on Knowledge
  Discovery \& Data Mining (KDD'18), pp. 2357--2366 (2018)

\bibitem{GAT}
Veli{\v{c}}kovi{\'{c}}, P., Cucurull, G., Casanova, A., Romero, A., Li{\`{o}},
  P., Bengio, Y.: {Graph Attention Networks}.
\newblock In: International Conference on Learning Representations (ICLR)
  (2018)

\bibitem{pointerNN}
Vinyals, O., Fortunato, M., Jaitly, N.: Pointer networks.
\newblock In: Advances in Neural Information Processing Systems 28, pp.
  2692--2700 (2015)

\bibitem{SDNE}
Wang, D., Cui, P., Zhu, W.: Structural deep network embedding.
\newblock In: Proceedings of ACM SIGKDD International Conference on Knowledge
  Discovery and Data Mining (KDD'16), pp. 1225--1234 (2016)

\bibitem{wang2019learning}
Wang, R., Yan, J., Yang, X.: Learning combinatorial embedding networks for deep
  graph matching.
\newblock In: Proceedings of the IEEE International Conference on Computer
  Vision (ICCV'19), pp. 3056--3065 (2019)

\bibitem{socialNetwork}
Wasserman, S., Faust, K., et~al.: Social network analysis: Methods and
  applications, vol.~8.
\newblock Cambridge university press (1994)

\bibitem{ImproveTSP}
Wu, Y., Song, W., Cao, Z., Zhang, J., Lim, A.: Learning improvement heuristics
  for solving the travelling salesman problem.
\newblock CoRR \textbf{abs/1912.05784} (2019)

\bibitem{WuSurvey}
Wu, Z., Pan, S., Chen, F., Long, G., Zhang, C., Yu, P.S.: A comprehensive
  survey on graph neural networks.
\newblock CoRR \textbf{abs/1901.00596} (2019)

\bibitem{dgk}
Yanardag, P., Vishwanathan, S.: Deep graph kernels.
\newblock In: Proceedings of ACM SIGKDD International Conference on Knowledge
  Discovery and Data Mining (KDD'15), pp. 1365--1374 (2015)

\bibitem{SPAGAN}
Yang, Y., Wang, X., Song, M., Yuan, J., Tao, D.: {SPAGAN}: Shortest path graph
  attention network.
\newblock In: Proceedings of International Joint Conference on Artificial
  Intelligence (IJCAI'19), pp. 4099--4105 (2019)

\bibitem{RUM}
{Yu}, Y., {Lu}, Z., {Liu}, J., {Zhao}, G., {Wen}, J.: {RUM}: Network
  representation learning using motifs.
\newblock In: IEEE International Conference on Data Engineering (ICDE'19), pp.
  1382--1393 (2019)

\bibitem{GEbio}
Yue, X., Wang, Z., Huang, J., Parthasarathy, S., Moosavinasab, S., Huang, Y.,
  Lin, S.M., Zhang, W., Zhang, P., Sun, H.: {Graph embedding on biomedical
  networks: methods, applications and evaluations}.
\newblock Bioinformatics \textbf{36}(4), 1241--1251 (2019)

\bibitem{ZhangSurvey}
{Zhang}, D., {Yin}, J., {Zhu}, X., {Zhang}, C.: Network representation
  learning: A survey.
\newblock IEEE Transactions on Big Data \textbf{6}(1), 3--28 (2020)

\bibitem{GE4Rec}
Zhang, F., Yuan, N.J., Lian, D., Xie, X., Ma, W.Y.: Collaborative knowledge
  base embedding for recommender systems.
\newblock In: Proceedings of ACM SIGKDD International Conference on Knowledge
  Discovery and Data Mining (KDD'16), pp. 353--362 (2016)

\end{thebibliography}


\end{document}